\documentclass[11pt]{article}

\usepackage[preprint]{acl}

\usepackage{times}
\usepackage{latexsym}

\usepackage[T1]{fontenc}

\usepackage[utf8]{inputenc}

\usepackage{microtype}

\usepackage{inconsolata}

\usepackage{graphicx}
\usepackage[most]{tcolorbox}
\usepackage{enumitem}
\usepackage{multirow}
\usepackage[table,xcdraw]{xcolor}
\usepackage{CJKutf8} 
\usepackage{fontawesome5}
\usepackage{xcolor, soul,todonotes}

\newcommand{\dscf}[0]{\textsc{CommunityFact}}

\title{\textsc{CommunityFact}: A Dynamic, Multilingual, Multi-domain Benchmark for Misinformation Detection in the Wild}

\author{Sahajpreet Singh ~~~~~ Insyirah Mujtahid ~~~~~ Min-Yen Kan ~~~~~ Kokil Jaidka \\
National University of Singapore \\
\texttt{sahajpreet.singh@u.nus.edu, \{insyirah, knmnyn, jaidka\}@nus.edu.sg}
}

\begin{document}
\maketitle
\begin{abstract}
Misinformation verification increasingly occurs in public, fast-moving, and multilingual online settings, where static benchmarks provide an incomplete measure of model reliability. We introduce \dscf{}, a refreshable benchmark for misinformation detection in the wild, with three major goals: coverage, granularity, and redistributability. This release contains 15,992 standalone claims across five languages and two domains. We evaluate ten LLMs under varying inference-time capabilities, including thinking and web-search. Our results show that closed-input verification remains challenging, web access yields the largest gains, and web-enabled LLMs' source-selection policies are systematically misaligned with the sources human Community Notes raters converge on --- a gap that closes through model-specific mechanisms of retrieval expansion or pruning. We further find substantial variation across language--domain slices and across the evidence ecosystems used by web-enabled systems. Beyond evaluation, \dscf{} positions Community Notes as a training signal for claim-conditioned source suggesters that could improve factual verification on novel claims.
\end{abstract}

\section{Introduction}
\label{sec:cf_intro}

{\let\thefootnote\relax\footnotetext{\textbf{Code:}
\href{https://github.com/sahajps/CommunityFact}{ \texttt{github.com/sahajps/CommunityFact}}}}

{\let\thefootnote\relax\footnotetext{\textbf{Data:}
\href{https://huggingface.co/datasets/sahajps/CommunityFact}{ \texttt{hf.co/datasets/sahajps/CommunityFact}}}}

{\let\thefootnote\relax\footnotetext{
\textbf{Disclaimer:} This paper contains examples of misleading content used solely for illustration purposes.}}

On X, users increasingly invoke Grok in public threads to adjudicate circulating claims, making the model's responses visible not only to the requester but to the broader conversation \cite{renault2026grok,mei2026grok}. In this setting, factual errors are not merely private hallucinations. They have become public verification failures that can amplify, legitimize, or prolong the spread of misinformation while undermining public trust \cite{li2025human}. Evaluating such deployed, web-enabled systems, therefore, requires benchmarks that reflect \textit{how} misinformation actually circulates online. However, existing factuality benchmarks largely frame claim verification as an offline task over fixed snapshots \cite{fatahi2025factbench}.

Benchmarks have long driven progress in AI \cite{blagec2021critical}, but misinformation in online environments is dynamic. Claims, evidence, and narratives shift temporally across events, languages, and communities \cite{su2020motivations,bessi2015trend,del2016spreading}. Using static benchmarks to evaluate LLM-based systems risks measuring memorization or contamination from widely circulated data rather than genuine verification ability \cite{sainz2023nlp,dong2024generalization,sun2025emperor}. Closing this gap requires benchmarks that are temporally fresh, granular enough to localize failure modes, and robust to the realities of social-media data and online narratives.

Recent dynamic factuality benchmarks, including Adv-Fake~\cite{chen2025real}, LiveFact~\cite{xu2026livefact}, and IRB~\cite{do2026irb}, partially address staleness by sourcing claims from recent web data. Three gaps remain for misinformation detection on social media. \textbf{(1) Coverage:} existing resources skew English-centric~\cite{wang2017liar,gupta2021x,barron2023clef} and draw from centralized outlets, encyclopedias, and professional fact-checkers~\cite{singh2024breaking,sahitaj2026articles}, underrepresenting localized and social-media narratives. \textbf{(2) Granularity:} post-level labels collapse multiple claims of mixed veracity into one verdict, hiding where verification fails and masking the systematic gap between true- and false-claim performance \cite{chen2022generating,glockner2024ambifc}.  \textbf{(3) Redistributability:} social-media posts erode under deletion, access restrictions, and copyright (e.g., X and PolitiFact data used in TripleFact~\cite{xu2025triplefact}). Therefore, we treat shareability as a design principle that the construction pipeline must satisfy by default, not a downstream caveat.

Community Notes\footnote{Community Notes is X's crowdsourced context feature, where volunteer contributors propose notes for potentially misleading posts; notes become publicly visible only after receiving sufficient helpfulness ratings from contributors with diverse viewpoints \cite{mohammadi2026birdwatch}.} offers a practical way to address these gaps because it provides a continuously updated stream of real-world claims, community-vetted corrective context, and evidence links attached to misinformation as it appears in the wild. Prior work has shown that Community Notes significantly reduce engagement with misleading posts, improve users' factual beliefs, and provide effective corrections at scale \cite{chuai2024did,slaughter2025community,vu2026trust}. These properties make it a particularly useful substrate for building benchmarks that are both grounded in naturally occurring misinformation and capable of being refreshed over time. 

At this point, we introduce \dscf{}, a dynamic, multilingual, multi-domain benchmark for real-world misinformation detection built from helpful Community Notes on X. Rather than releasing raw tweet--note pairs, \dscf{} uses them as a construction source. We filter helpful notes, pair them with source post metadata, and convert them into standalone factual claims with note-grounded binary labels. The evaluated release contains \textbf{15{,}992} claim-level examples across five languages---English, Spanish, French, Japanese, and Portuguese---and two domains, \textsc{Politics} and \textsc{Finance}. Each released example includes the claim, timestamp, language, domain, label, and evidence URLs; the original post and note text are not required for benchmark use. Because the pipeline is parameterized by the Community Notes archive snapshot and date range, the benchmark can be regenerated for future archive releases fully automatically. We discuss the detailed benchmark construction and its human validation in Section \ref{sec:cf_data_cons}.

Additionally, \dscf{} supports evaluation along three axes that are rarely studied together in prior misinformation benchmarks: \textbf{(1)} temporal generalization through recent holdout examples within each language--domain group, \textbf{(2)} multilingual and multi-domain analysis under a shared construction process, and \textbf{(3)} capability-stratified evaluation of LLMs under instruction-only, reasoning-enabled, web-search, and evidence-guided web-search settings. Because aggregate scores mask large cross-slice variation in multilingual misinformation verification, our work treats per-slice reporting as a first-class evaluation requirement, not an optional breakdown. The web-search settings are especially important because LLMs are increasingly used as public, ad-hoc verification systems \cite{augenstein2024factuality}. 

We find that closed-input verification trails web-enabled verification at every model scale; that web-enabled LLMs systematically diverge from the sources Community Notes contributors cite and raters perceive as helpful; and that aligning retrieval toward these human-vetted sources improves verification through model-specific expansion or pruning, motivating future work on claim-conditioned source suggesters.

\section{Related Work}
\label{sec:cf_rel_work}
\paragraph{Misinformation and factuality benchmarks.}
Misinformation detection has often been framed as supervised verification over fixed collections of claims, news articles, Wikipedia evidence, or social-media posts \cite{thorne2018fever,shu2020fakenewsnet,ostrowski2021multi,gupta2021x,wang2023check,shafiei2025multihoax}. These resources remain important, but they differ substantially in temporal scope, language coverage, domain focus, and dependence on platform-specific snapshots, making them imperfect proxies for recent social-media misinformation \cite{xu2024benchmark}. Recent dynamic factuality benchmarks address temporal limitations through newer web content, temporal splits, or refreshable construction pipelines \cite{lin2025fact,altuncu2025factors,do2026irb,xu2026livefact}. \dscf{} follows this dynamic evaluation direction, but targets misleading social-media claims and narratives that may not be covered well by mainstream media due to coverage limitations or even selection bias \cite{singh2024independent}.

\paragraph{Community Notes and evidence-grounded verification.}
Community Notes provide crowd-sourced corrections to public posts and have been studied for note helpfulness or automatic note generation \cite{xing2026communitynotes,singh2026gitsearch,zhang2025commenotes,singh2025limitations}. Prior work in this line treats Community Notes primarily as signals to analyze, predict, or generate. Ours uses Community Notes differently: rather than predicting note helpfulness or generating notes, it converts helpful tweet--note pairs into standalone factual claims with note-grounded labels and evidence URLs.

\section{The \dscf{} Benchmark}
\label{sec:cf_data_cons}
\dscf{} is a binary misinformation classification benchmark constructed from X's Community Notes. This release contains 15{,}992 standalone claims across five languages and two domains. The dataset contains 9{,}970 \textsc{True} claims and 6{,}022 \textsc{False} claims. Each example includes a claim, note-grounded label, timestamp, language, domain, and evidence URLs. This section describes the construction pipeline, label and evidence retrieval, temporal split, and human validation.

\paragraph{Construction pipeline and release setup.}
The construction pipeline is guided by three design goals: examples should be text-verifiable without attached media, shareable without requiring users to reconstruct raw social-media posts, and analyzable across language, domain, and temporal slices. We start from the public Community Notes release\footnote{\url{https://x.com/i/communitynotes/download-data}}, using the notes along with their helpfulness status history. For this version of the benchmark, we use the archive snapshot containing notes contributions through March 27, 2026, and restrict the source notes to those created in 2025. This approximately three-month lag between note creation and the archive cutoff allows note ratings to stabilize before inclusion \cite{chuai2026consensus}. 

\begin{figure}[!ht]
    \centering
    \includegraphics[width=\linewidth]{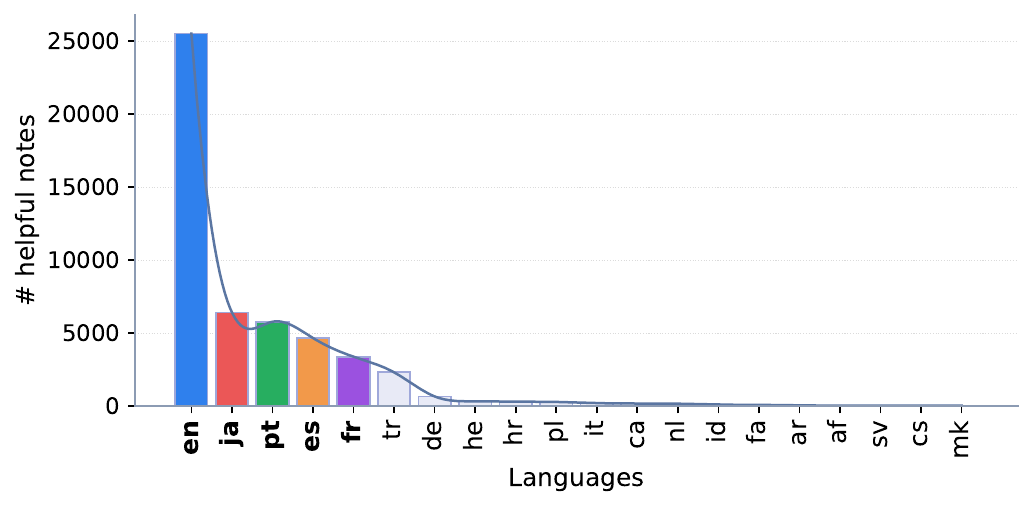}
    \caption{Language distribution of helpful Community Notes before top-k language benchmark filtering.}
    \label{fig:cf_crhn_lang_dist}
    \vspace{-\baselineskip}
\end{figure}

Within this window, we retain notes whose helpfulness status is \texttt{CURRENTLY\_RATED\_HELPFUL}, keep only the most recent helpful note for each source post, and remove media notes so that examples can be verified from text alone. We then detect note language and select the five most frequent helpful non-media note languages: English, Japanese, Portuguese, Spanish, and French. These languages cover 90.1\% of helpful non-media notes, while the remaining languages form a long tail, as shown in Figure~\ref{fig:cf_crhn_lang_dist}. To support domain-level analysis, we classify each note as \textsc{Politics}, \textsc{Finance}, or \textsc{Other} with GPT-5.5\footnote{\label{fn:cf_gpt5.5}\texttt{gpt-5.5-2026-04-23}} using the prompt in Appendix~\ref{app:cf_claim-prompt}, retaining the \textsc{Politics} and \textsc{Finance} subsets. We retrieve post metadata through the X API and keep only post--note pairs for which metadata is available and the post language matches the note language to keep cross-lingual ambiguity minimum, yielding 11{,}545 source post--note pairs for claim construction.

\paragraph{Claim, label, and evidence construction.}
For each retained post--note pair, the builder extracts a number of checkworthy, self-contained factual claims using GPT-5.5$^{\text{\ref{fn:cf_gpt5.5}}}$ with \texttt{reasoning\_effort = medium}. The extraction prompt requires each claim to contain the entities, dates, locations, quantities, jurisdictions, and other qualifiers needed for verification without access to the original post or note. It also excludes claims that depend on attached media, account authenticity, post availability, or other platform metadata. Labels are grounded in the helpful Community Note; a claim is labeled \textsc{False} when it captures a factual assertion from the post that the note refutes or corrects, and \textsc{True} when it captures a factual assertion supported or clarified by the note. Later, a self-refinement step audits each generated claim for standalone completeness, note support, label correctness, scope validity, and non-duplication. The respective prompt templates can be found in Appendix \ref{app:cf_claim-prompt}. We also preserve URLs cited in the helpful note. These URLs are treated as community-selected evidence leads rather than oracle proofs and are used only in the evidence-guided evaluation setting.

\paragraph{Release format and temporal split.}
The public release contains standalone claim-level examples with the claim, note-grounded label, timestamp, language, domain, and evidence URLs; raw post and note text are retained only in the private construction version for auditing. The release is produced by a parameterized filtering and construction pipeline. Given a new Community Notes archive snapshot and a specified date range, the same pipeline can be rerun to generate an updated benchmark release.

For the evaluation in this paper, one run of the pipeline yields a 2025 release with 15{,}992 examples. We split examples temporally within each domain--language group, holding out the most recent examples for testing while assigning all claims derived from the same post--note pair to the same split. This prevents source--pair leakage and evaluates models on temporally later misinformation narratives. The final split contains 12{,}414 training examples and 3{,}578 test examples; detailed label counts by split, domain, and language are reported in Table~\ref{tab:cf_split-distribution}. Representative benchmark examples are shown in Appendix~\ref{app:cf_data_ex}.

\begin{table}[!h]
\centering
\scriptsize
\setlength{\tabcolsep}{3.5pt}
\renewcommand{\arraystretch}{1.12}
\resizebox{\columnwidth}{!}{
\begin{tabular}{llccccc}
\hline
\textbf{Split} & \textbf{Domain} & \textbf{en} & \textbf{es} & \textbf{fr} & \textbf{ja} & \textbf{pt} \\
\hline
\multirow{2}{*}{Full}
& Politics & 3753/2530 & 1073/672 & 908/516 & 1571/878 & 694/464 \\
& Finance  & 996/506   & 209/125  & 165/99  & 197/121  & 404/111 \\
\hline
\multirow{2}{*}{Train}
& Politics & 3435/2347 & 762/482 & 591/333 & 1252/697 & 391/266 \\
& Finance  & 685/316   & 127/73  & 97/61   & 117/74   & 267/41 \\
\hline
\multirow{2}{*}{Test}
& Politics & 318/183 & 311/190 & 317/183 & 319/181 & 303/198 \\
& Finance  & 311/190 & 82/52   & 68/38   & 80/47   & 137/70 \\
\hline
\end{tabular}
}
\caption{
\dscf{} train--test distribution by domain and language.
Each cell reports \textsc{True}/\textsc{False} claim counts.}
\label{tab:cf_split-distribution}
\vspace{-\baselineskip}
\end{table}

\paragraph{Human validation.}
After automatic construction, we validate \dscf{} using a primary in-house audit and an external Prolific\footnote{\url{https://www.prolific.com/}} audit; the full protocol is detailed in Appendix~\ref{app:cf_he_guide}. We recruit Prolific annotators through a sample-based screening process (native speakers from respective countries) and collect 5.3 annotations per example on average, reporting majority votes over examples with at least three annotations. 

The in-house audit indicates high construction quality across validation criteria, with near-ceiling agreement on claim completeness, checkability, note--claim relevance, and label consistency, and strong agreement on domain assignment (cf. Table~\ref{tab:cf_human_validation} in Appendix~\ref{app:cf_he_guide}). The Prolific audit provides an independent robustness check for criteria that can be reliably assessed by non-expert annotators. Majority votes remain high for checkability and note--claim relevance, while label consistency is somewhat lower but still strong at 90.9\%. For independent timestamped factuality judgments, we rely on the in-house audit\footnote{We do not use crowd-worker factuality judgments as a primary validation metric because this task requires careful, temporally grounded fact-checking of multilingual real-world claims, which is substantially more demanding and requires professional-level fact-checking experience.}, where note-grounded labels agree with factuality judgments for 90.3\% of examples. Together, these results suggest that \dscf{} is both faithful to helpful Community Notes and broadly aligned with independent factuality judgments.

Later, inspecting the 9.7\% of examples where note-grounded labels diverge from independent factuality judgments, we observe three recurring patterns. \textit{(i) Extraction drift:} the standalone claim occasionally sharpens or specifies the source assertion (e.g., ``the highest'' softened to ``among the highest''), shifting what is being verified relative to what the note refutes. \textit{(ii) Interpretive scope:} some claims hinge on contested legal, historical, or geopolitical framings where the note and an independent judge reasonably adopt different framings of the same fact. E.g., the status of Kashmir, Manchukuo, or the UK's nuclear deterrent, etc.  \textit{(iii) Indirect evidence:} a note sometimes addresses the surrounding context rather than the specific assertion, leaving room for stronger inferences than the cited evidence strictly supports. These patterns reflect the inherent difficulty of compressing nuanced, real-world claims into binary verdicts and motivate future extensions to richer label taxonomies (cf. Limitations).

\begin{table*}[t]
\centering
\resizebox{\textwidth}{!}{
\begin{tabular}{llccccclcccccc}
\hline
\multirow{2}{*}{\textbf{\begin{tabular}[c]{@{}l@{}}Model\\ Capabilities\end{tabular}}} & \multirow{2}{*}{\textbf{Model}} & \multicolumn{5}{c}{\textbf{Politics}} &  & \multicolumn{5}{c}{\textbf{Finance}} & \multirow{2}{*}{\textbf{Overall}} \\ \cline{3-7} \cline{9-13}
 &  & en & es & fr & ja & pt &  & en & es & fr & ja & pt &  \\ \hline
Instruction & Aya-Expanse-8B & 62.71 & 61.1 & 61.67 & 69.63 & 62.47 &  & 59.35 & 65.21 & 65.52 & 74.97 & 65.5 & 63.74 \\
Following & Ministral-8B & 39.59 & 36.79 & 37.28 & 35.04 & 34.56 &  & 37.54 & 36.92 & 31.49 & 34.44 & 38.64 & 36.74 \\
 & EuroLLM-9B & 47.06 & 40.6 & 45.57 & 52.43 & 47.72 &  & 45.12 & 43.19 & 39.06 & 48.22 & 49.44 & 46.46 \\
 & EuroLLM-22B & 54.75 & 56.5 & 56.17 & 71.21 & 59.53 &  & 51.94 & 56.09 & 64.15 & 66.88 & 60.36 & 59.23 \\
 & Aya-Expanse-32B & 71.79 & 69.37 & 66.04 & 74.9 & 66.71 &  & 69.7 & 69.89 & 59.72 & 79.13 & 70.95 & 69.93 \\ \hline
Reasoning & Qwen3-14B & 62.07 & 62.07 & 59.8 & 70.4 & 63.72 &  & 59.6 & 61.94 & 53.77 & 64.53 & 63.14 & 62.83 \\
 & ~~+Thinking & 56.99 & 54.8 & 56.39 & 67.0 & 57.9 &  & 57.79 & 59.26 & 56.46 & 62.84 & 62.31 & 58.92 \\ \cline{2-14} 
 & Qwen3-32B & 68.63 & 65.15 & 62.7 & 66.8 & 64.83 &  & 64.4 & 65.73 & 62.21 & 73.03 & 65.36 & 65.68 \\
 & ~~+Thinking & 62.65 & 57.75 & 60.56 & 62.19 & 60.72 &  & 61.09 & 60.18 & 60.38 & 66.11 & 62.71 & 61.15 \\ \hline
Web-search & GPT-5-nano & 39.45 & 36.79 & 37.01 & 38.08 & 37.39 &  & 38.16 & 34.48 & 35.01 & 40.35 & 43.21 & 38.09 \\
 & ~~+Thinking & 55.47 & 50.89 & 51.62 & 57.54 & 55.35 &  & 57.77 & 51.29 & 52.56 & 61.8 & 53.14 & 54.81 \\
 & ~~~~+Web-search & 81.2 & 74.83 & 73.32 & 79.23 & 76.21 &  & 73.95 & 69.98 & 59.45 & 83.62 & 69.42 & 75.59 \\ 
\cline{2-14} 
 & Gemini-2.5-Flash & 65.66 & 67.2 & 67.49 & 74.17 & 66.06 &  & 65.45 & 64.17 & 62.46 & 81.6 & 68.01 & 68.01 \\
 & ~~+Thinking & 67.45 & 65.6 & 65.71 & 70.74 & 65.67 &  & 63.58 & 63.38 & 63.51 & 78.48 & 69.0 & 66.97 \\
 & ~~~~+Web-search & 82.37 & 78.06 & 75.57 & 82.33 & 76.55 &  & \multicolumn{1}{l}{75.94} & 72.94 & 68.18 & 92.37 & 72.76 & 78.13 \\ 
\cline{2-14} 
 & Grok-4.3 & 54.93 & 51.7 & 52.73 & 58.83 & 53.06 &  & 54.0 & 50.33 & 51.73 & 52.11 & 52.07 & 53.85 \\
 & ~~+Thinking & 67.65 & 66.82 & 70.72 & 77.07 & 73.24 &  & 69.63 & 72.37 & 76.69 & 78.48 & 73.3 & 71.58 \\
 & ~~~~+Web-search & \textbf{86.09} & \textbf{80.23} & \textbf{82.45} & \textbf{88.51} & \textbf{83.04} &  & \textbf{83.1} & \textbf{81.76} & \textbf{77.02} & \textbf{92.49} & \textbf{81.65} & \textbf{83.8} \\  
 \hline
\end{tabular}
}
\caption{Model performance on \dscf{} using macro-F1 scores on the temporally held-out test split, overall and by domain--language slice for \textsc{Politics} and \textsc{Finance} across English (en), Spanish (es), French (fr), Japanese (ja), and Portuguese (pt). Models are grouped by inference capability: instruction-following, reasoning-enhanced, and web-search settings. \texttt{+Thinking} denotes explicit reasoning, and \texttt{+Web-search} denotes web verification. Higher scores indicate better factuality classification performance.}
\label{tab:cf_main_results}
\vspace{-\baselineskip}
\end{table*}

\section{Experimental Setup}
\label{sec:exp_setup}
In this section, we discuss models employed for benchmarking and their capabilities, prompting methods, and evaluation strategy.

\paragraph{Models and inference settings.}
We benchmark three capability tiers of multilingual LLMs and evaluate each under the inference modes it supports. 
\textbf{(1)} \textit{Instruction-following models} include Aya-Expanse-8B, Aya-Expanse-32B \cite{dang2024ayaexpanse}, Ministral-8B \cite{ministral8b2410repo}, EuroLLM-9B \cite{martins2025eurollm9btechnicalreport}, and EuroLLM-22B \cite{ramos2026eurollm22btechnicalreport}. These models are evaluated in a closed-input setting, receiving only the claim and timestamp, and serve as the multilingual factuality baseline. 
\textbf{(2)} \textit{Reasoning-capable open models} include Qwen3-14B and Qwen3-32B \cite{qwen3technicalreport}. We run them both with and without explicit thinking, while withholding external evidence in both cases, to isolate the effect of inference-time reasoning. 
\textbf{(3)} \textit{Web-enabled models} include GPT-5-nano \cite{singh2026openaigpt5card}, Gemini-2.5-Flash \cite{comanici2025gemini25pushingfrontier}, and Grok-4.3 \cite{grok43docs2026}. These systems are evaluated under direct-answer, thinking, unguided web-search, and evidence-guided web-search conditions. In open web-search, the model independently retrieves timestamp-relevant evidence; in evidence-guided web-search, we additionally provide the URLs cited by the helpful Community Note as community-selected leads rather than oracle justifications. Models may use these URLs when relevant or search beyond them when they are inaccessible, incomplete, stale, or insufficient. 
This design compares parametric knowledge, explicit reasoning, open-web retrieval, and Community-Note-guided evidence under a shared claim-verification format. Model identifiers and capability settings are reported in Appendix~\ref{app:cf_model_det}.

\paragraph{Prompting.}
We evaluate \dscf{} under a zero-shot prompting regime. All claims are evaluated in their original language, without translation or task-specific fine-tuning. All systems use a unified verdict-oriented prompt template. The prompt specifies the claim, timestamp, and available evidence condition, and asks the model to return a final \textsc{True} or \textsc{False} label. For web-search settings, the prompt emphasizes evidence-grounded verification rather than plausibility-based answering. For evidence-guided web-search, note-provided URLs are presented as prioritized sources while preserving access to general web search. Full prompt templates are provided in Appendix~\ref{app:cf_bench_prompts}. For open-weight models, decoding is deterministic; API-based systems use the corresponding provider settings, with reasoning and web-search tools enabled only in the conditions that require them.

\paragraph{Metrics.}
The primary metric is macro-F1 over the \textsc{True} and \textsc{False} classes. Macro-F1 equally weights both labels and is therefore appropriate for \dscf{}, where class balance varies across languages and domains. We report macro-F1 overall and for each domain--language slice.

\begin{figure*}[t]
    \centering
    \includegraphics[width=\linewidth]{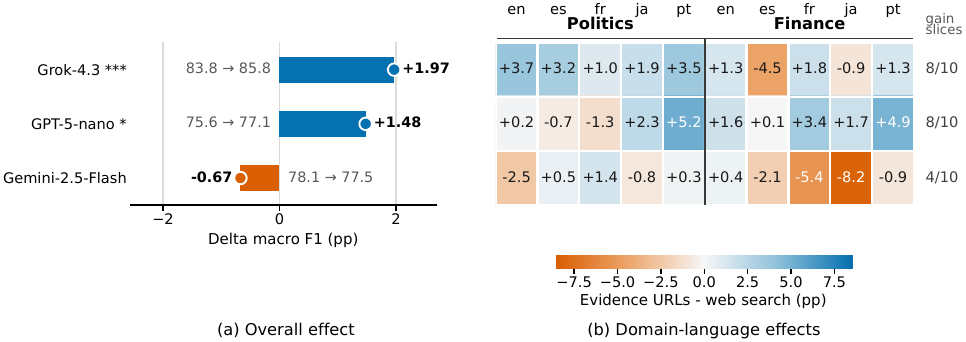}
    \caption{Effects of evidence-guided web verification. Panel \textbf{(a)} reports the overall macro-F1 change, in pp, when web-enabled models are given evidence URLs cited by helpful Community Notes in addition to web search. Significance stars denote paired significance tests on the overall evaluation set only: *$ p<0.05$, **$p<0.01$, ***$p<0.001$. Panel \textbf{(b)} shows descriptive domain-language effects, where blue indicates gains and orange indicates declines relative to unguided web search. The “gain slices” column counts positive domain-language slices. Detailed per-slice scores and deltas are provided in Appendix \ref{app:cf_evigui_scores}.}
    \label{fig:cf_evi_eff}
    \vspace{-\baselineskip}
\end{figure*}

\section{Results \& Discussion}
\label{sec:cf_res}
We organize the discussion around two key research questions: \vspace{0.7em} \\ 
\textbf{RQ1:} \textit{How do LLMs with different inference-time capabilities perform on temporally held-out misinformation claims across languages and domains?}\\
\textbf{RQ2:} \textit{Does evidence-guided web-search reshape the evidence ecosystems of web-enabled LLMs, and does alignment with human-curated sources improve verification beyond default web-search?}

\subsection{Inference-time Capabilities}
Closed-input verification is hard. In Table~\ref{tab:cf_main_results}, Aya-Expanse-32B leads instruction-following models at 69.93 macro-F1; same-class Aya-Expanse-8B reaches 63.74, while Ministral-8B trails at 36.74. \textbf{Scale alone does not buy verification ability.} Multilingual instruction tuning and factual calibration dominate parameter count, since \dscf{} test claims sit temporally later than training and often concern recent or localized events.

If parameters alone do not close the gap, a natural next lever is inference-time reasoning, but explicit reasoning is unreliable without evidence. \texttt{+Thinking} hurts both Qwen3 variants (62.83$\rightarrow$58.92; 65.68$\rightarrow$61.15) and Gemini-2.5-Flash (68.01$\rightarrow$66.97), yet lifts Grok-4.3 (53.85$\rightarrow$71.58) and GPT-5-nano (38.09$\rightarrow$54.81). Reasoning mode is therefore \emph{not} a universal lever; deploying it as a default for fact-checking can silently degrade verification depending on the base model, and the gains it does provide remain bounded by what the model already knows.

Grounding reasoning in retrieved evidence, by contrast, produces the largest and most consistent gains. Web access raises GPT-5-nano to 75.59, Gemini-2.5-Flash to 78.13, and Grok-4.3 to 83.80 macro-F1, with Grok-4.3 leading every domain--language slice. Yet slice variance persists --- GPT-5-nano, for example, scores 59.45 on French finance against 83.62 on Japanese finance. Here, two implications follow. First, every closed-input model, regardless of size or reasoning mode, sits categorically below the weakest web-enabled one. Production fact-checkers without retrieval are below a floor that frontier closed-input models cannot reach. Second, any benchmark that reports only aggregate F1 leaves systematic, language-specific failures invisible.

\begin{figure}[!h]
    \centering
    \includegraphics[width=\linewidth]{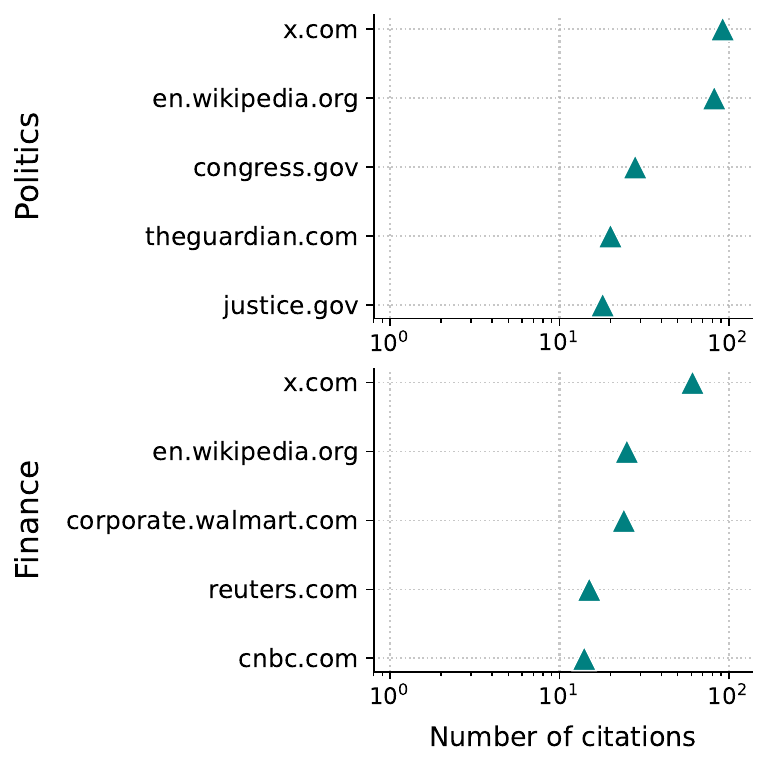}
    \caption{Observed source-domain ecosystems for crowd-sourced verification (\textit{via} Community Notes). Here, we plot the most frequently emitted source domains for the English subset of \dscf{} across domains. The x-axis is logarithmic.}
    \label{fig:cf_h_top_domain_en}
    \vspace{-\baselineskip}
\end{figure}

\begin{figure*}[!t]
    \centering
    \includegraphics[width=\linewidth]{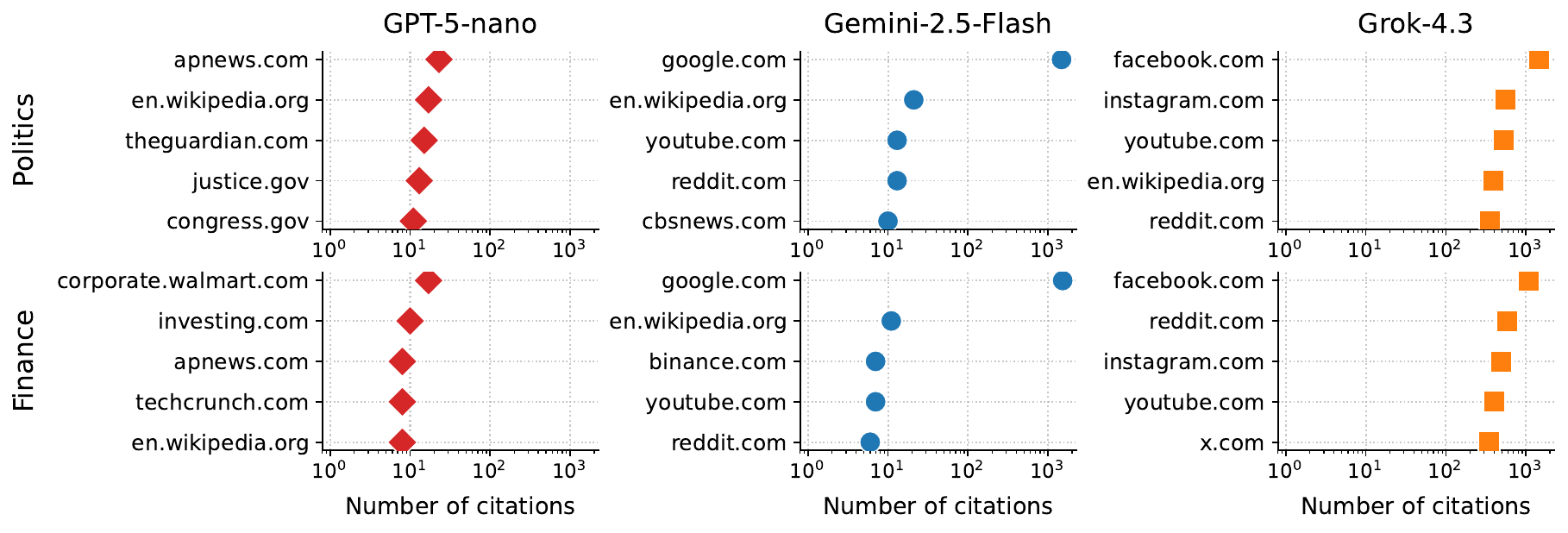}
    \caption*{(a) Open web-search setting}
    \includegraphics[width=\linewidth]{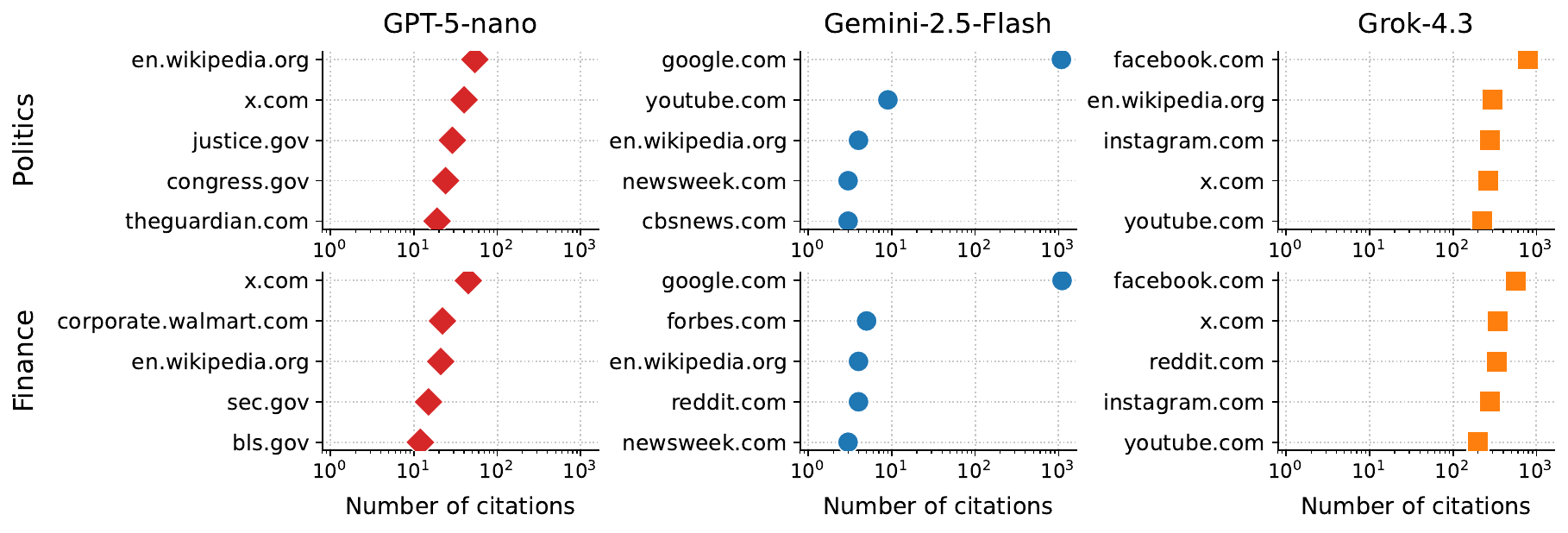}
    \caption*{(b) Evidence-guided web-search setting}
    \caption{Observed source-domain ecosystems for web-search verification. For each web-enabled model, we plot the most frequently emitted source domains for the English subset of \dscf{} across domains (cf. Appendix \ref{app:cf_url_det} for other languages). The x-axis is logarithmic.}
    \label{fig:cf_top_domains_en}
    \vspace{-\baselineskip}
\end{figure*}

\subsection{Does Evidence-guided Search Help?}
Yes for two of three web-enabled systems, and the asymmetry reveals \emph{how} models integrate community-surfaced evidence rather than merely \emph{whether} more evidence helps. Note-cited URLs significantly improve GPT-5-nano ($+1.48$; $p$<$0.05$) and Grok-4.3 ($+1.97$; $p$<$0.001$) under paired tests (Figure~\ref{fig:cf_evi_eff}, Appendix~\ref{app:cf_evigui_scores}), while Gemini-2.5-Flash drops $0.67$ pp (not significant). The two gainers improve in 8 of 10 domain--language slices, with the largest gains on Portuguese politics ($+5.24$) and finance ($+4.86$) for GPT-5-nano. Slice-level reversals persist --- Grok-4.3 declines on Spanish finance ($-4.47$), and Gemini-2.5-Flash regresses sharply on Japanese finance ($-8.21$) --- showing that note URLs act as leads, not oracles.

\paragraph{Alignment with human-preferred helpful sources drives the gains.}
Figure~\ref{fig:cf_h_top_domain_en} shows the source-domain ecosystem actually used by Community Notes contributors when writing notes on English claims (cf. Appendix \ref{app:cf_url_det} for other languages), which were later found helpful by raters. These include authoritative institutional and journalistic outlets such as \texttt{justice.gov}, \texttt{congress.gov}, and \texttt{theguardian.com} for politics, and \texttt{cnbc.com}, \texttt{reuters.com}, and \texttt{corporate.walmart.com} for finance, alongside cross-cutting references to \texttt{en.wikipedia.org} and \texttt{x.com}. This is the human reference distribution against which we measure each model's retrieval. Table~\ref{tab:cf_hit_ratio} reports the corresponding domain hit-ratio. Under guided search, overlap with the domains present in the helpful Community Notes rises sharply for Grok-4.3 (0.62 to 0.94) and GPT-5-nano ($+0.43$), but is flat for Gemini-2.5-Flash ($-0.02$). The two systems that improve in macro-F1 are exactly the two that successfully realign their retrieval toward the human-vetted source set; the system that does not realign does not improve. Performance gains come from \emph{which} sources are consulted relative to the crowd reference, not from being handed more sources to consult.

\begin{table}[!h]
\resizebox{\columnwidth}{!}{
\begin{tabular}{lccc}
\hline
\textbf{Model} & \textbf{\begin{tabular}[c]{@{}c@{}}Open \\ web-search\end{tabular}} & \textbf{\begin{tabular}[c]{@{}c@{}}Evidence-guided \\ web-search\end{tabular}} & $\Delta$ \\ \hline
Grok-4.3 & 0.62 & 0.94 & {\color[HTML]{009901} +0.32} \\
GPT-5-nano & 0.13 & 0.56 & {\color[HTML]{009901} +0.43} \\
Gemini-2.5-Flash & 0.04 & 0.02 & {\color[HTML]{FE0000} -0.02} \\ \hline
\end{tabular}}
\caption{Domain hit-ratio in different web-search settings with respect to crowd-sourced fact-checking.}
\label{tab:cf_hit_ratio}
\vspace{-\baselineskip}
\end{table}

\paragraph{Gainers follow opposite efficiency profiles.}
The two improving systems converge on the human reference from opposite directions (Appendix \ref{app:cf_url_det}). GPT-5-nano \emph{expands} its retrieval (1.0$\rightarrow$1.4 URLs/prediction; source coverage 53.1\%$\rightarrow$73.1\%), using note leads to populate a previously sparse evidence trail that undershot the Community Notes citation density of $\sim$1.9 URLs/note. Grok-4.3 \emph{contracts} its retrieval (21.9$\rightarrow$13.9 URLs/prediction) while improving accuracy. A $\sim$36\% reduction in citation volume for a $+1.97$ pp macro-F1 gain, i.e., better verification at lower retrieval cost, pruning away from its over-broad social-platform default toward something closer to the human reference. Community-curated evidence thus functions as a focusing prior in both directions: it lets undercitation grow toward useful evidence and lets overcitation prune away noise.

\paragraph{Why Gemini does not benefit.}
Gemini-2.5-Flash's open-web ecosystem is dominated by generic google search and aggregator endpoints (\texttt{google.com/search}, \texttt{youtube.com}, \texttt{reddit.com}; Figure~\ref{fig:cf_top_domains_en}a) that barely intersect the human reference distribution in Figure~\ref{fig:cf_h_top_domain_en} --- baseline overlap of 0.04. Under guided search, its citation count falls (3.8$\rightarrow$2.5 URLs) and source coverage drops (99.4\%$\rightarrow$75.4\%), yet the source mix does not shift toward Community-Note-cited domains (Figure~\ref{fig:cf_top_domains_en}b; hit ratio unchanged). The model reduces retrieval effort without redirecting it; on slices like Japanese finance ($-8.21$), this actively hurts because the generic search it abandons was the channel surfacing locale-relevant evidence. The broader picture across web-enabled systems makes the gap concrete: citation density spans an order of magnitude (1.0--21.9 URLs/prediction), and locale further shapes provenance, with GPT-5-nano routing Japanese politics through \texttt{factcheckcenter.jp} and \texttt{mhlw.go.jp}, Grok-4.3 through \texttt{sankei.com} and \texttt{nikkei.com}, but Gemini through generic \texttt{google.com} and \texttt{ja.wikipedia.org} (Appendix~\ref{app:cf_url_det}). Note URLs --- drawn from the authoritative outlets, government records, and locale-specific endpoints visible in Figure~\ref{fig:cf_h_top_domain_en} --- integrate cleanly with GPT-5-nano's conventional retrieval and Grok-4.3's broader mix, but conflict with a generic-search-anchored default. Integrating community-surfaced leads is therefore itself a model capability that must be measured separately from web access.

\paragraph{Implications.}
These results should not be read as a recipe to wait for a Community Note before adjudicating a claim. Once a helpful note exists, the verification is in large part already done. Rather, they establish that \textbf{web-enabled LLMs' source-selection policies are systematically misaligned with the sources humans converge on}, and that closing this gap improves verification in two characteristic ways: expanding sparse retrieval toward human-vetted sources (GPT-5-nano) and pruning over-broad retrieval to a more focused, human-aligned set (Grok-4.3). The relevant follow-up direction is therefore not note-conditioned inference, but \emph{learning from notes}. Community Notes supply a continuous, multilingual stream of $\langle$\emph{claim}, \emph{sources humans found verifiable}$\rangle$ pairs from which one can train \textbf{claim-conditioned source suggesters} --- lightweight retrieval priors that, given a fresh claim, propose the kind of sources a human verifier would consult. Plugged into a web-enabled LLM, such a suggester inherits the alignment advantage we measure here while remaining deployable on novel claims for which no note yet exists; \dscf{} quantifies the headroom available to this direction and diagnoses whether a given verifier needs an expansion or a pruning prior to benefit.

\section{Conclusion}
\label{cf:conc}
We introduced \dscf{}, a refreshable benchmark for misinformation verification in the wild, built from helpful Community Notes on X. The 2025 release contains 15{,}992 standalone, claim-level examples across five languages and two high-impact domains, with note-grounded labels, timestamps, and community-cited evidence URLs, while avoiding dependence on raw social-media text. Its parameterized construction pipeline enables future releases from new Community Notes snapshots.

Our evaluation across ten models and four inference settings shows that closed-input verification remains difficult and that web access provides the largest gains. We further find that web-enabled LLMs' source-selection policies are systematically misaligned with the sources human contributors converge on, and that closing this gap improves verification through retrieval expansion in some models and retrieval pruning in others. Aggregate scores alone are therefore insufficient, since performance varies sharply across language and domain slices and web-enabled systems rely on distinct evidence ecosystems with different auditability profiles. Beyond evaluation, \dscf{} positions Community Notes as a training signal rather than only an inference-time aid. Future work can use these traces to learn claim-conditioned source suggesters that propose human-aligned sources for fresh claims, extend the benchmark to further languages and domains, and incorporate multimodal claims as Community Notes coverage of images and video continues to grow.

\section*{Limitations}
While \dscf{} provides a scalable and multilingual benchmark for evaluating factual verification in dynamic social-media settings, several design choices delimit the scope of the current release. We discuss these limitations below, together with the steps taken to mitigate them and the directions planned for future extensions.

\paragraph{Coverage.}
The benchmark reflects the distribution of claims and languages available in the public Community Notes archive. As a result, coverage is naturally affected by the long-tail distribution of language use and by variation in Community Notes activity across regions, topics, and communities. Rather than treating this as a fixed constraint, our pipeline is designed to be continuously extensible: as additional languages and claim types become better represented in the archive, they can be incorporated automatically into future benchmark versions. The current release focuses on two misinformation-related domains and is intentionally text-only. Future releases will expand the topical scope to additional domains such as science, entertainment, etc, and may also incorporate multimodal content where reliable provenance and annotation signals are available.

\paragraph{Label provenance.}
The labels in \dscf{} are grounded in helpful Community Notes rather than independent oracle judgments. This choice enables scalable, transparent, and dynamically updatable benchmark construction, but it also introduces potential label noise. To assess this risk, we conducted an in-house audit and observed 90.3\% agreement with independent factuality verdicts. The remaining disagreement indicates that some claims may be ambiguous, context-dependent, or imperfectly captured by the helpful note. In addition, helpful-note selection is shaped by X's bridge-based rating algorithm and by the demographics and participation patterns of Community Notes contributors, which may underrepresent some regions, languages, or viewpoints. These factors should be considered when interpreting benchmark results. At the same time, grounding labels in helpful notes allows \dscf{} to support a fully automated and dynamic benchmark-generation pipeline, which is central to our goal of tracking factual verification performance over time.

\paragraph{Binary labels.}
\dscf{} operationalizes verification as a binary \textsc{True}/\textsc{False} decision aligned with the verdict implied by the helpful note. This formulation supports clear evaluation and direct comparison across models, but it necessarily abstracts away more fine-grained factuality distinctions. For example, claims that are misleading without being strictly false, partially correct, dependent on missing context, or unverifiable from available evidence are not explicitly modeled in the current version. Extending the benchmark to include richer label taxonomies is an important direction for future work.

\paragraph{Web-search reproducibility.}
Results for web-enabled systems depend on factors that may change over time, including each provider's live index, query routing, retrieval policy, geographic context, and the accessibility of note-cited URLs (e.g., login, paywalls, \texttt{robots.txt}, etc.). Consequently, absolute scores for web-enabled systems should be interpreted as point-in-time estimates rather than immutable properties of the underlying model. To support reproducibility and follow-up analysis, we snapshot the backend metadata produced during web search, including search queries, retrieved sources, and reasoning traces whenever applicable. We will release these logs together with the code and dataset repository, enabling researchers to inspect retrieval behavior, reproduce our analysis more closely, and study web-augmented factual verification without incurring the full cost of large-scale proprietary web-search calls. This is particularly important for groups with limited resources, since web-search APIs can cost tens of USD per 1K sources.

\section*{Ethical Statement}
We adhere to the ethical guidelines for the use of social media data and AI-assisted dataset construction. All data underlying \dscf{} was obtained from publicly available sources, such as X's Community Notes data archive and the X API for post metadata. The released benchmark contains paraphrased, standalone claims rather than raw post or note text, which reduces re-identification risk and avoids the redistributability pitfalls of raw social-media corpora. The dataset includes real-world political and financial claims describing sensitive or contested events; these are retained solely to support the evaluation of misinformation-detection systems. This study and its human evaluation protocol received an exemption from our Departmental Ethics Review Committee (DERC) under the Institutional Review Board (IRB), given the use of publicly available data and minimal risk to participants. Prolific annotators were compensated according to the platform’s recommended fair pay rate. Before participating in the survey, participants were also required to provide consent for the use of the data they submitted by selecting a mandatory checkbox to proceed.

We acknowledge that any misinformation benchmark could be repurposed to fine-tune systems that generate more persuasive misinformation. We mitigate this by (a) releasing only note-grounded claim-label pairs and community-cited evidence URLs rather than original misleading posts, and (b) framing \dscf{} as an evaluation resource for verification systems intended to augment human moderators rather than replace them.

\section*{Acknowledgments}
This research was supported by the Ministry of Education, Singapore, through the AcRF TIER 3 Grant (MOE-MOET32022-0001). We sincerely thank the human evaluators for their time and valuable assistance in this work. We would also like to thank Lyle Ungar (University of Pennsylvania) for his insightful feedback on the initial idea.

\bibliography{custom}

\appendix
\section{Experimental Details}
\label{app:cf_exp_det}

\subsection{Computational Resources}
\label{app:cf_comp_res}
Inference for all open-source models was conducted on a commodity GPU cluster comprising 8× NVIDIA H100 GPUs, with identical hyperparameter settings applied across models to ensure consistency.

\subsection{Dataset Construction Pipeline Prompts}
\label{app:cf_claim-prompt}
We use three prompts in the benchmark construction pipeline: domain classification, claim-label extraction, and claim-label refinement, given in Figures \ref{fig:cf_domain_class_prompt}, \ref{fig:cf_claim_ext_prompt}, and \ref{fig:cf_self_ref_prompt}, respectively.

\subsection{Model Details}
\label{app:cf_model_det}
The different models employed in the experiments are listed in Table \ref{tab:model_details}, along with their corresponding model-card names. The evaluated models were obtained either as publicly available checkpoints from Hugging Face or accessed through external APIs such as the OpenAI API. For open models, we set \texttt{do\_sample = False} to ensure reproducible outputs. For closed models, we use the default settings, enable the web-search tool, and set the \texttt{reasoning\_effort = "medium"} whenever applicable.

\begin{table}[!h]
\centering
\resizebox{\columnwidth}{!}{
\begin{tabular}{llll}
\hline
\textbf{Model Name} & \textbf{Detailed Model Name} & \faBrain & \faSearch \\ \hline
Aya-Expanse-8B & CohereLabs/aya-expanse-8b & \textcolor{red}{\faTimes} & \textcolor{red}{\faTimes} \\
Ministral-8B & mistralai/Ministral-8B-Instruct-2410 & \textcolor{red}{\faTimes} & \textcolor{red}{\faTimes} \\
EuroLLM-9B & utter-project/EuroLLM-9B-Instruct-2512 & \textcolor{red}{\faTimes} & \textcolor{red}{\faTimes} \\
EuroLLM-22B & utter-project/EuroLLM-22B-Instruct-2512 & \textcolor{red}{\faTimes} & \textcolor{red}{\faTimes} \\
Aya-Expanse-32B & CohereLabs/aya-expanse-32b & \textcolor{red}{\faTimes} & \textcolor{red}{\faTimes} \\
Qwen3-14B & Qwen/Qwen3-14B & \textcolor{green}{\faCheck} & \textcolor{red}{\faTimes} \\
Qwen3-32B & Qwen/Qwen3-32B & \textcolor{green}{\faCheck} & \textcolor{red}{\faTimes} \\
GPT-5-nano & gpt-5-nano-2025-08-07 & \textcolor{green}{\faCheck} & \textcolor{green}{\faCheck} \\
Gemini-2.5-Flash & gemini-2.5-flash & \textcolor{green}{\faCheck} & \textcolor{green}{\faCheck} \\
Grok-4.3 & grok-4.3 & \textcolor{green}{\faCheck} & \textcolor{green}{\faCheck} \\ \hline
\end{tabular}}
\caption{Models evaluated in \dscf{} experiments and their corresponding model-card or API names. The final two columns indicate (yes: \textcolor{green}{\faCheck} or no: \textcolor{red}{\faTimes}) whether each model is evaluated with explicit reasoning (\faBrain) and web-search (\faSearch) capabilities.}
\label{tab:model_details}
\vspace{-\baselineskip}
\end{table}

\subsection{Benchmarking Prompts}
\label{app:cf_bench_prompts}
For the different inference settings we studied, we present the prompt templates used in Figures \ref{fig:cf_zeroshot_prompt}, \ref{fig:cf_websearch_prompt}, and \ref{fig:cf_evi_websearch_prompt} for the zero-shot (and reasoning), web-search, and guided web-search settings, respectively.

\section{Human Validation Protocol}
\label{app:cf_he_guide}
We conduct two complementary validation studies for \dscf{}: an in-house expert audit and an external crowd audit on Prolific. Both studies evaluate whether automatically constructed examples are valid benchmark instances, but they serve different roles. Because \dscf{} labels are derived from helpful Community Notes, the primary question is whether the extracted claim and label are faithful to the post--note pair. We therefore treat the in-house audit as the primary validation signal, and use Prolific as an external robustness check.

\paragraph{In-house audit.}
The in-house audit covers 361 examples. Annotators are shown the original post, helpful Community Note, extracted claim, assigned domain, assigned \textsc{True}/\textsc{False} label, timestamp, and URLs cited by the note. For non-English examples, we additionally provide English translations of the relevant post, note, and claim text. This setup reduces language-comprehension noise and allows annotators to focus on the benchmark-specific validity questions: whether the claim is self-contained, checkable, addressed by the note, assigned to the correct domain, and labeled consistently with the note.

\paragraph{Prolific audit.}
We also collect external annotations through Prolific. Annotators are recruited using a sample-based screening process, and each example is assigned to multiple workers. The resulting audit contains 5.3 annotations per example on average. To reduce individual crowd noise, we aggregate labels by majority vote and report results over the 309 examples with at least three annotations. Because crowd annotators vary in background and familiarity with note-grounded labeling, we interpret Prolific results as an external robustness signal rather than a replacement for the in-house audit.

\paragraph{Annotation questions.}
Annotators answer whether the extracted claim is self-contained, whether it is checkable, whether the assigned domain is correct, whether the helpful Community Note directly addresses the claim, whether the assigned label is consistent with the note, and whether the note-cited URLs provide direct, partial, weak, or no evidence for the claim. We also ask a separate independent factuality question: whether the claim is true, false, or has insufficient evidence at the provided timestamp. This separates construction fidelity from real-world factuality: an example is valid for \dscf{} when the claim and label faithfully reflect the helpful Community Note, while the factuality audit measures whether this note-grounded label also agrees with independent verification. We present the detailed guidelines provided to human subjects in Figure \ref{fig:cf_he_guide}.

We summarize the results of our human evaluation in Table \ref{tab:cf_human_validation}.

\begin{table}[!h]
\centering
\scriptsize
\setlength{\tabcolsep}{4pt}
\renewcommand{\arraystretch}{1.08}
\resizebox{\columnwidth}{!}{
\begin{tabular}{lcc}
\hline
\textbf{Validation criterion} & \textbf{In-house} & \textbf{Prolific} \\
\hline
Self-contained claim & 99.7 & 84.1 \\
Checkable claim & 100.0 & 96.4 \\
Correct domain & 96.7 & 89.0 \\
Note addresses claim & 100.0 & 96.1 \\
Label consistent with note & 100.0 & 90.9 \\
Evidence quality (direct+partial support) & 85.9 & 94.2 \\
Label matches factuality audit & 90.3 & --- \\
\hline
\end{tabular}
}
\caption{Human validation of \dscf{} examples. 
In-house results are computed over 361 expert-audited examples and serve as the primary validation signal. Prolific results are majority votes over 309 examples with at least three independent annotations.}
\label{tab:cf_human_validation}
\vspace{-\baselineskip}
\end{table}

\section{Representative Examples}
\label{app:cf_data_ex}
Table \ref{tab:cf_data_ex} showcases a sample from our benchmark dataset, \dscf{}.

\section{Effect of Guided Search}
\label{app:cf_evigui_scores}
Table \ref{tab:cf_evigui_scores} provides the detailed scores corresponding to Figure \ref{fig:cf_evi_eff}.

\section{Details on Cited URLs}
\label{app:cf_url_det}
We present the average number of URLs used by Community Notes writers and web-search-enabled LLMs during fact-checking in Table \ref{tab:cf_cite_stats}. The same table also reports the percentage of the test set for which at least one URL source was used.

\begin{figure*}[!t]
    \includegraphics[width=0.99\linewidth]{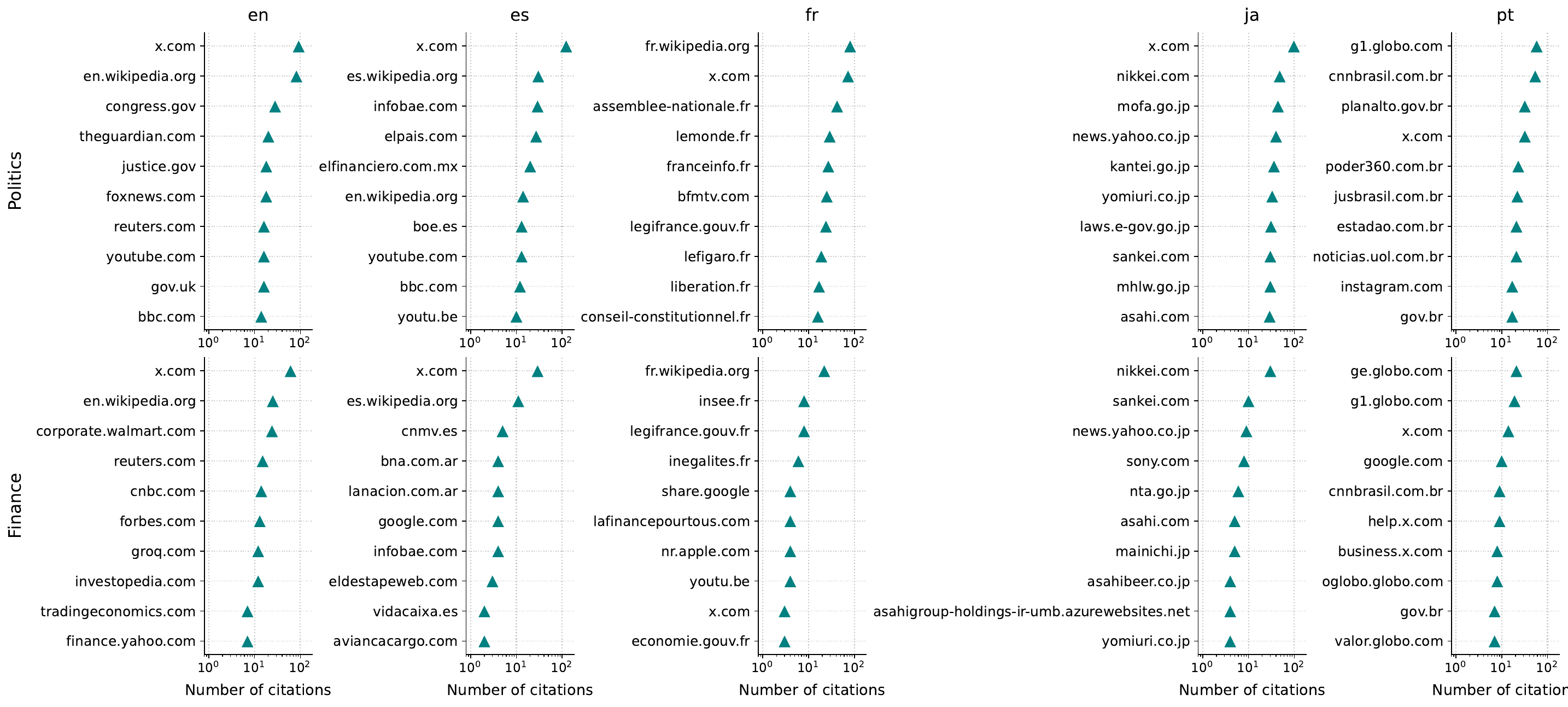}
    \caption{Observed top-10 source-domain ecosystems for crowd-sourced verification (\textit{via} Community Notes) across domain-language slices. The x-axis is logarithmic.}
    \label{fig:cf_h_top_domain}
\end{figure*}

\begin{figure*}[!t]
    \includegraphics[width=0.99\linewidth]{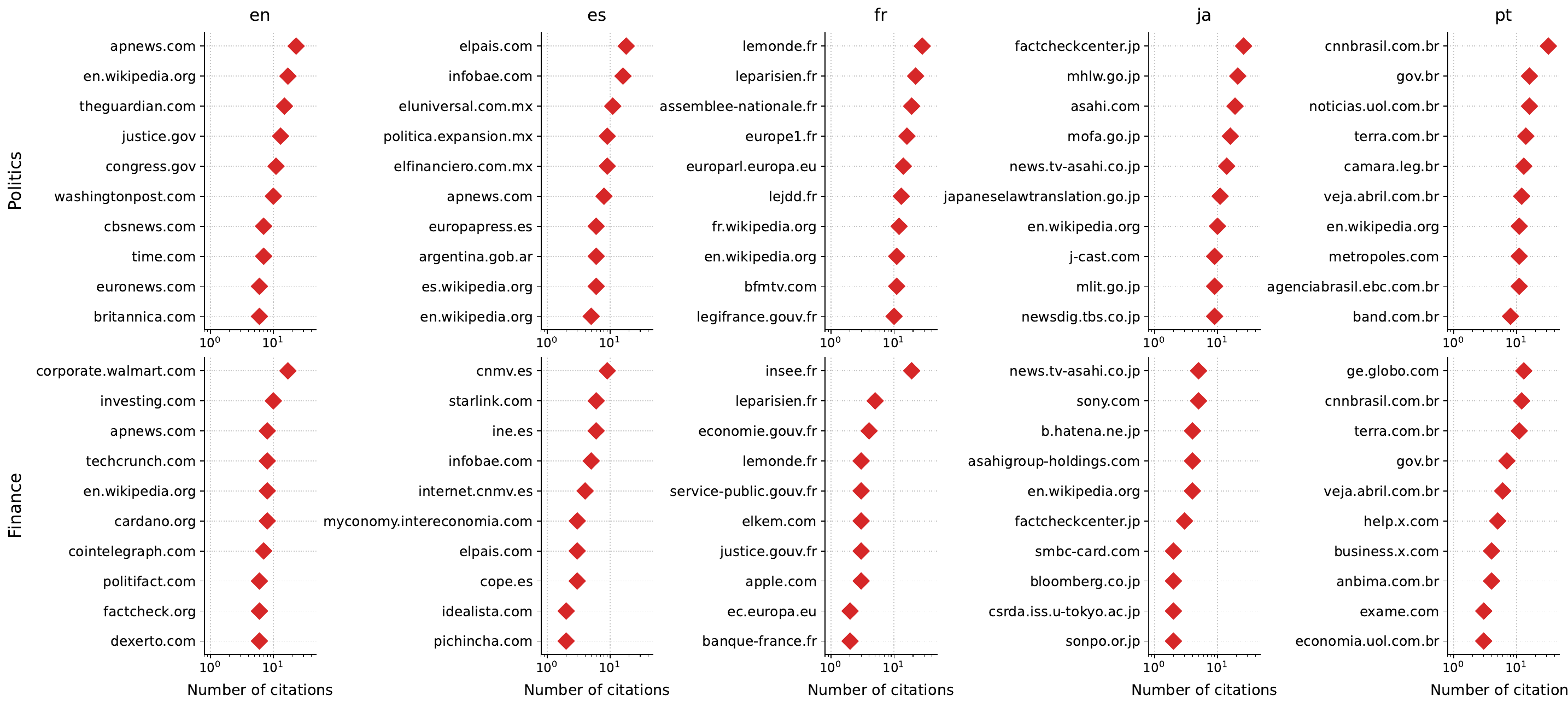}
    \caption*{(a) GPT-5-nano}
    \includegraphics[width=0.99\linewidth]{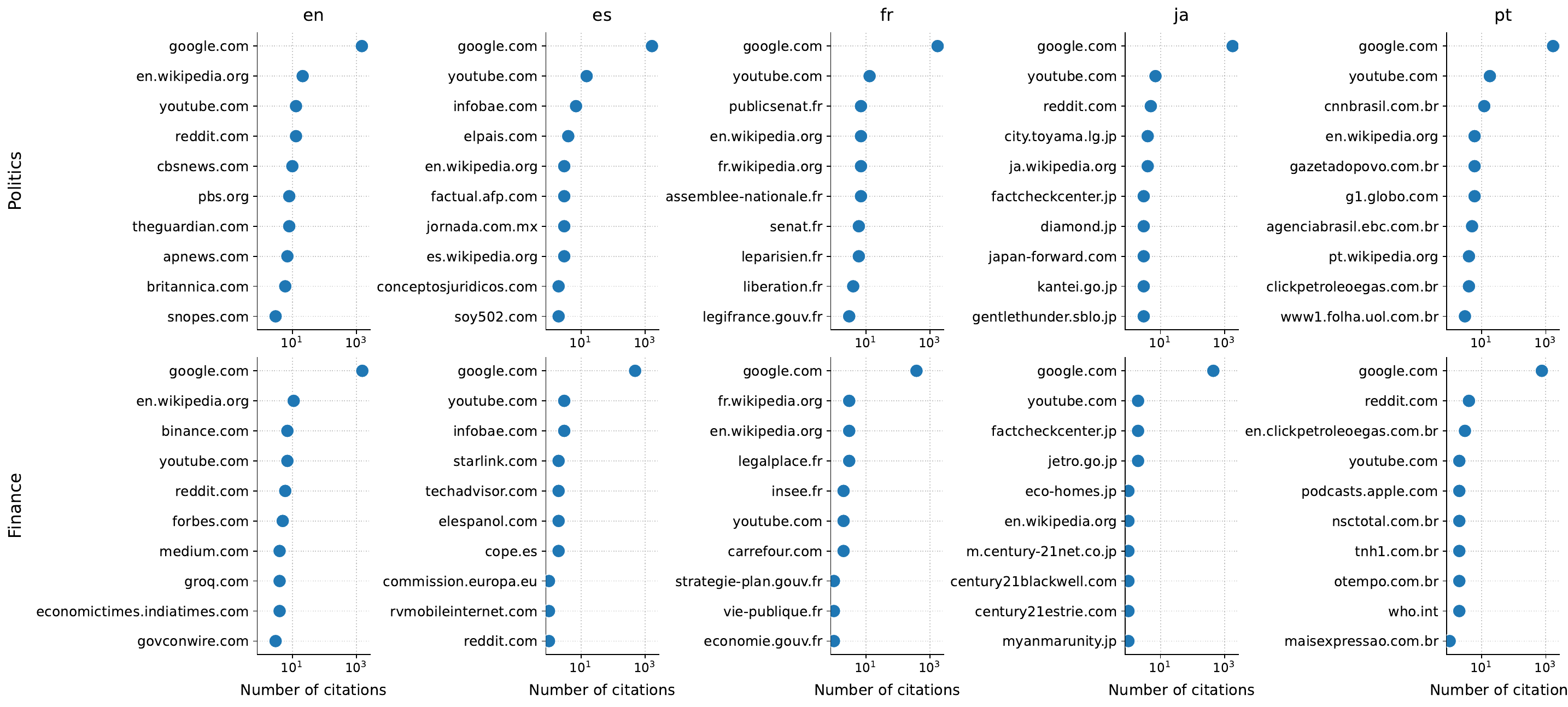}
    \caption*{(b) Gemini-2.5-Flash}
    \includegraphics[width=0.99\linewidth]{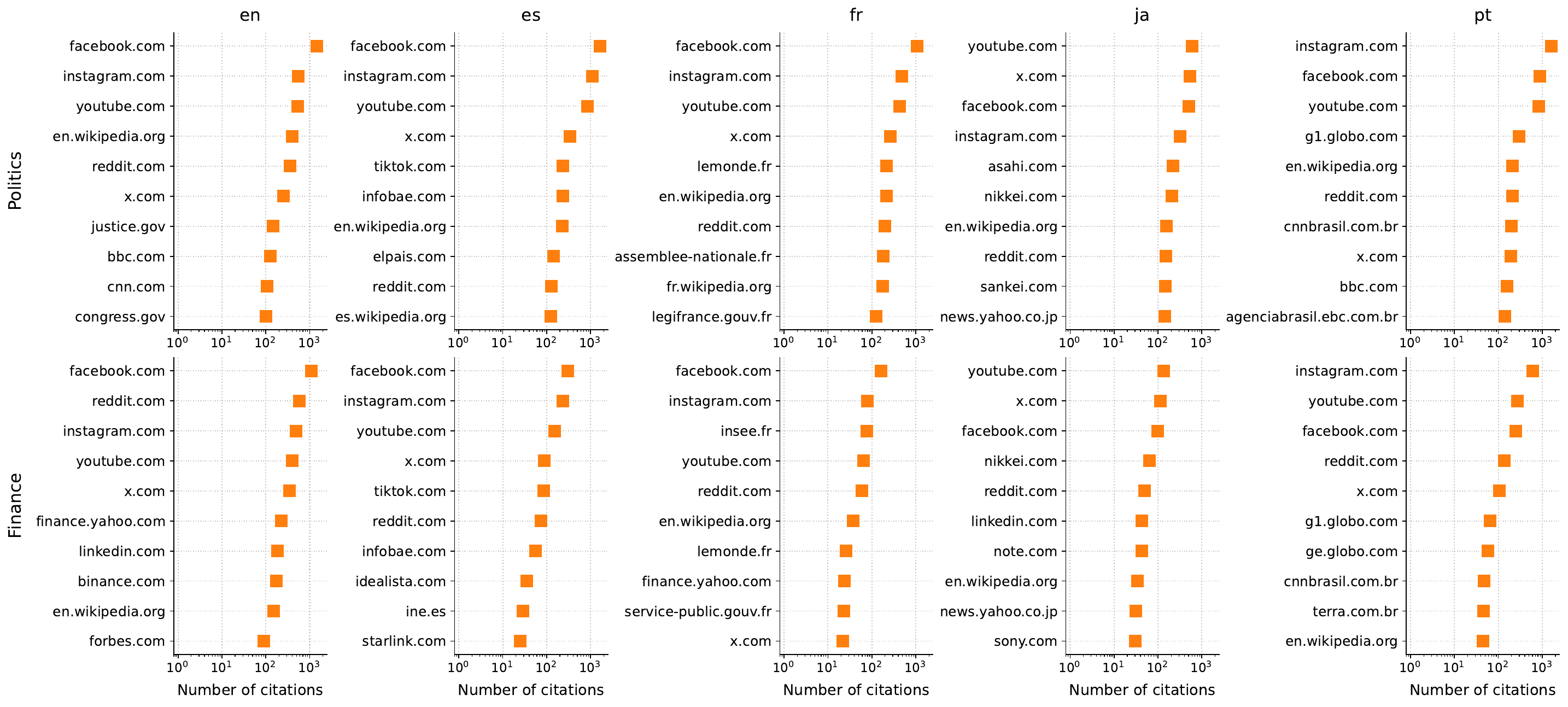}
    \caption*{(c) Grok-4.3}
    \caption{Observed top-10 source-domain ecosystems for web-search verification across domain-language slices. The x-axis is logarithmic.}
    \label{fig:cf_top_domain}
\end{figure*}

\begin{figure*}[!t]
    \includegraphics[width=0.99\linewidth]{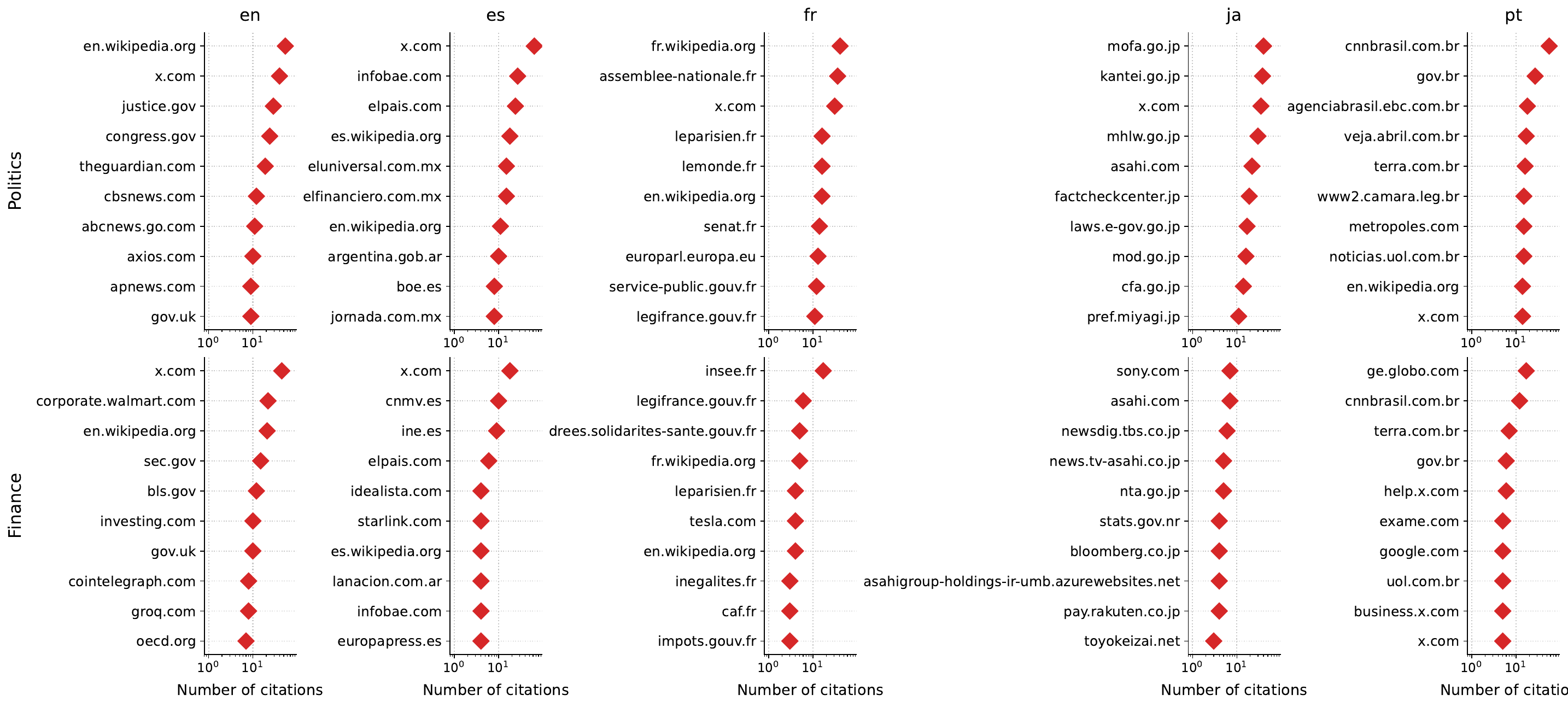}
    \caption*{(a) GPT-5-nano}
    \includegraphics[width=0.99\linewidth]{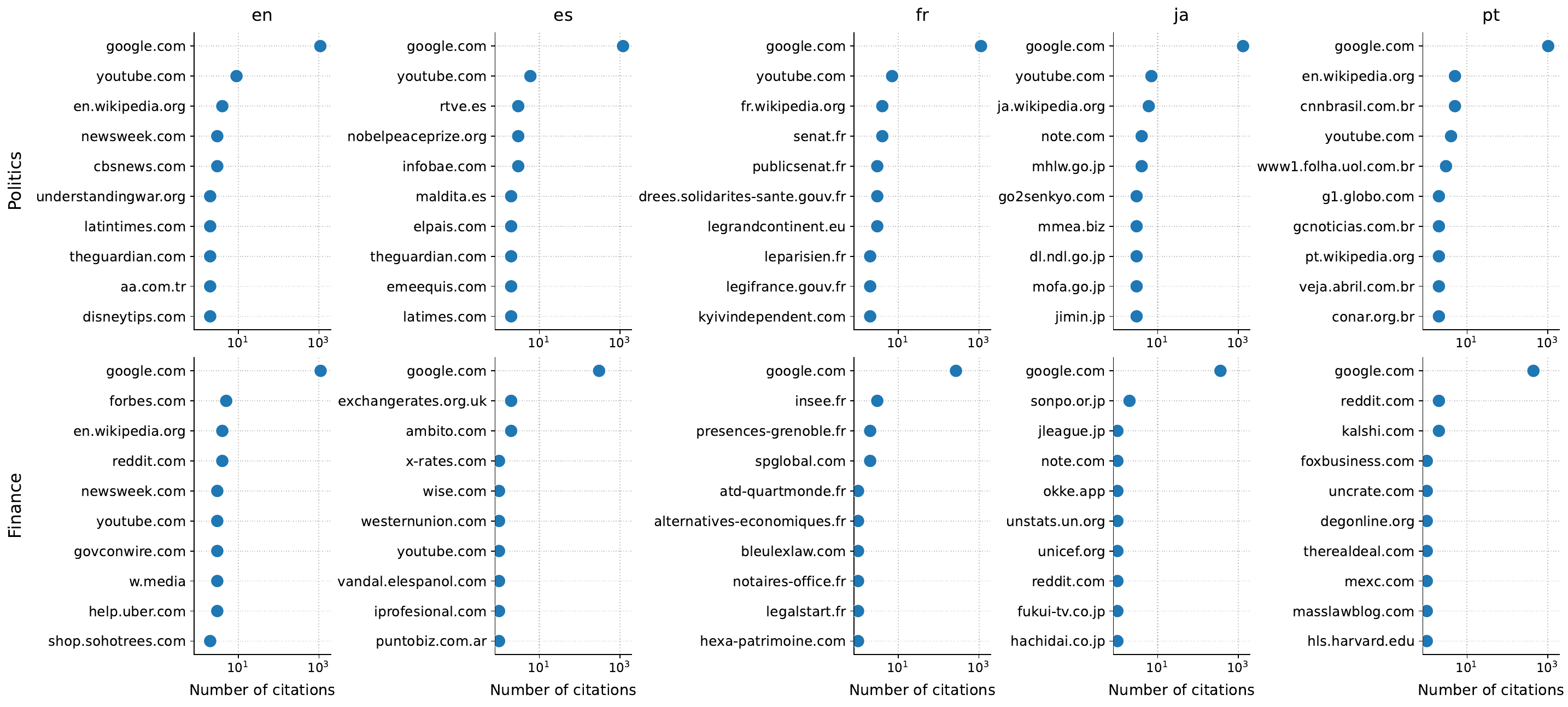}
    \caption*{(b) Gemini-2.5-Flash}
    \includegraphics[width=0.99\linewidth]{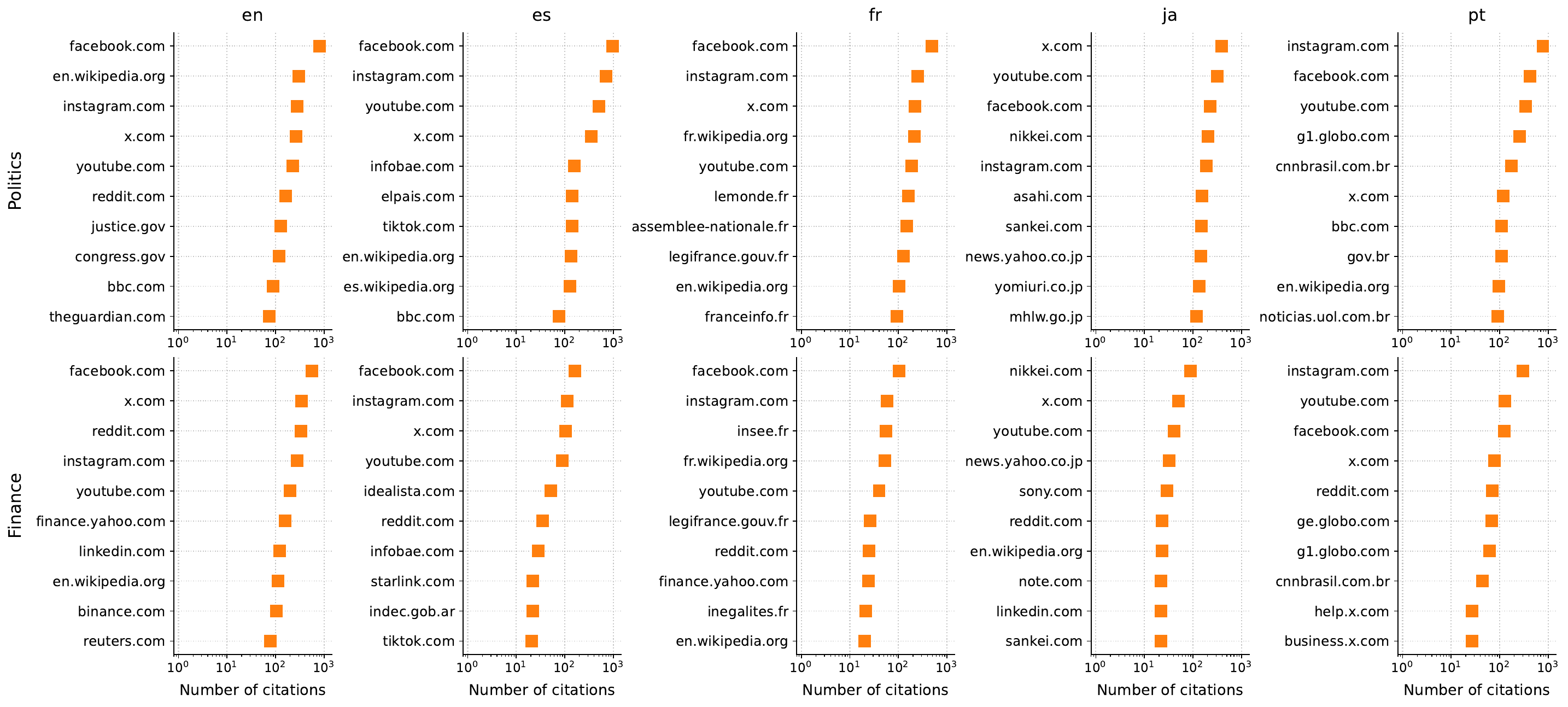}
    \caption*{(c) Grok-4.3}
    \caption{Observed top-10 source-domain ecosystems for evidence-guided web-search verification across domain-language slices. The x-axis is logarithmic.}
    \label{fig:cf_gs_top_domain}
\end{figure*}

Figure \ref{fig:cf_h_top_domain} extends the English-only view in Figure \ref{fig:cf_h_top_domain_en} to the rest of the languages. Additionally, Figures \ref{fig:cf_top_domain} and \ref{fig:cf_gs_top_domain} extend Figures \ref{fig:cf_top_domains_en}(a) and \ref{fig:cf_top_domains_en}(b), respectively, from the English subset to all language–domain slices.

\section{Cost and Reusability}
\label{app:cf_cost_reuse}
\dscf{} required substantial monetary investment to construct, validate, and evaluate. The current release cost approximately USD 800 for GPT-5.5-based domain classification, claim extraction, label construction, and refinement. Multilingual human validation (on Prolific) costs more than EUR 500. Large-scale experiments with proprietary reasoning and web-search-enabled LLMs added costs on the order of thousands of USD. 

We also incurred paid API costs to retrieve source-post, a.k.a. tweet metadata. These costs reflect a growing barrier to building reusable social-media benchmarks under current platform policies. We therefore designed \dscf{} to be reusable and reshared. The released benchmark does not require raw post or note text for evaluation, and includes standalone claims, labels, timestamps, domains, languages, and evidence URLs. Our goal is to let the community study dynamic, web-grounded misinformation verification without requiring each to absorb the same construction and validation costs.

\begin{table*}[!t]
\resizebox{\linewidth}{!}{
\begin{tabular}{p{1.2\columnwidth}cccc}
\hline
\textbf{Claim} & \textbf{Language} & \textbf{Domain} & \textbf{Label} & \textbf{Timestamp} \\ \hline
55 million people are living in the United States on U.S. visas. & en & Politics & False & Aug 22, 2025  \\ \hline
UK consumers can submit car finance mis-selling claims themselves for free. & en & Finance & True & Feb 01, 2025  \\ \hline
Apple desactivó temporalmente la traducción en vivo de los AirPods en la Unión Europea por cumplimiento de la Ley de Mercados Digitales. & es & Politics & True & Sep 13, 2025  \\
\textbf{English Translation:} Apple temporarily disabled live translation on AirPods in the European Union to comply with the Digital Markets Act. &  &  &  &   \\ \hline
Tebas percibe un salario anual total superior a 5.000.000 euros, un 1471\% más que en 2013. & es & Finance & True & Feb 28, 2025  \\
\textbf{English Translation:} Tebas receives a total annual salary exceeding 5,000,000 euros, 1471\% more than in 2013. &  &  &  &   \\ \hline
En France, les fonctionnaires perdent un jour de congé ou un jour de RTT au titre de la journée de solidarité, comme les salariés du privé. & fr & Politics & True & Jun 06, 2025  \\
\textbf{English Translation:} In France, civil servants lose a day of leave or a day of RTT (reduced working time) as part of the solidarity day, just like private sector employees. &  &  &  &   \\ \hline
Les auto-entrepreneurs en France ne cotisent pas pour leur retraite ni pour la sécurité sociale. & fr & Finance & False & Oct 25, 2025  \\
\textbf{English Translation:} Self-employed individuals in France do not contribute to their retirement or social security. &  &  &  &   \\ \hline
\begin{CJK}{UTF8}{ipxm}米海軍はベネズエラ領空を侵犯した。\end{CJK} & ja & Politics & False & Dec 10, 2025  \\
\textbf{English Translation:} The U.S. Navy violated Venezuelan airspace. &  &  &  &   \\ \hline
\begin{CJK}{UTF8}{ipxm}日本の農業は第一次産業に分類されない。\end{CJK} & ja & Finance & False & Apr 06, 2025  \\
\textbf{English Translation:} Agriculture in Japan is not classified as a primary industry. &  &  &  &   \\ \hline
Um processo judicial no Brasil envolve um usuário que perdeu acesso a uma conta do WhatsApp, alegou prejuízos comerciais e obteve decisão judicial para restabelecer essa conta. & pt & Politics & True & Feb 12, 2025  \\
\textbf{English Translation:} A lawsuit in Brazil involves a user who lost access to a WhatsApp account, claimed commercial losses, and obtained a court order to reinstate that account. &  &  &  &   \\ \hline
A Associação Brasileira de Psiquiatria nomeou Felca recentemente como parceiro nacional da saúde mental. & pt & Finance & False & Aug 18, 2025 \\
\textbf{English Translation:} The Brazilian Psychiatric Association recently named Felca as a national partner for mental health. &  &  &  &   \\ \hline
\end{tabular}}
\caption{Representative examples from \dscf{}.}
\label{tab:cf_data_ex}
\end{table*}

\begin{table*}[!h]
\resizebox{\linewidth}{!}{
\begin{tabular}{llcccccccccccc}
\hline
 &  & \multicolumn{5}{c}{\textbf{Politics}} & \multicolumn{1}{l}{} & \multicolumn{5}{c}{\textbf{Finance}} &  \\ \cline{3-7} \cline{9-13}
\multirow{-2}{*}{\textbf{Model}} & \multirow{-2}{*}{\textbf{Type}} & en & es & fr & ja & pt & \multicolumn{1}{l}{} & en & es & fr & ja & pt & \multirow{-2}{*}{\textbf{Overall}} \\ \hline
GPT-5-nano & Web-search & 81.2 & 74.83 & 73.32 & 79.23 & 76.21 &  & 73.95 & 69.98 & 59.45 & 83.62 & 69.42 & 75.59 \\
 & ~~~+Evidences & 81.45 & 74.17 & 72.01 & 81.5 & 81.45 &  & 75.56 & 70.04 & 62.83 & 85.35 & 74.28 & 77.07 \\
 & \cellcolor[HTML]{EFEFEF}$\Delta$ & \cellcolor[HTML]{EFEFEF}{\color[HTML]{009901} 0.25} & \cellcolor[HTML]{EFEFEF}{\color[HTML]{FE0000} -0.66} & \cellcolor[HTML]{EFEFEF}{\color[HTML]{FE0000} -1.31} & \cellcolor[HTML]{EFEFEF}{\color[HTML]{009901} 2.27} & \cellcolor[HTML]{EFEFEF}{\color[HTML]{009901} 5.24} & \cellcolor[HTML]{EFEFEF} & \cellcolor[HTML]{EFEFEF}{\color[HTML]{009901} 1.61} & \cellcolor[HTML]{EFEFEF}{\color[HTML]{009901} 0.06} & \cellcolor[HTML]{EFEFEF}{\color[HTML]{009901} 3.38} & \cellcolor[HTML]{EFEFEF}{\color[HTML]{009901} 1.73} & \cellcolor[HTML]{EFEFEF}{\color[HTML]{009901} 4.86} & \cellcolor[HTML]{EFEFEF}{\color[HTML]{009901} 1.48} \\ \hline
Gemini-2.5-Flash & Web-search & 82.37 & 78.06 & 75.57 & 82.33 & 76.55 &  & 75.94 & 72.94 & 68.18 & 92.37 & 72.76 & 78.13 \\
 & ~~~+Evidences & 79.89 & 78.53 & 77.01 & 81.51 & 76.85 &  & 76.32 & 70.86 & 62.83 & 84.16 & 71.87 & 77.46 \\
 & \cellcolor[HTML]{EFEFEF}$\Delta$ & \cellcolor[HTML]{EFEFEF}{\color[HTML]{FE0000} -2.48} & \cellcolor[HTML]{EFEFEF}{\color[HTML]{009901} 0.47} & \cellcolor[HTML]{EFEFEF}{\color[HTML]{009901} 1.44} & \cellcolor[HTML]{EFEFEF}{\color[HTML]{FE0000} -0.82} & \cellcolor[HTML]{EFEFEF}{\color[HTML]{009901} 0.3} & \cellcolor[HTML]{EFEFEF} & \cellcolor[HTML]{EFEFEF}{\color[HTML]{009901} 0.38} & \cellcolor[HTML]{EFEFEF}{\color[HTML]{FE0000} -2.08} & \cellcolor[HTML]{EFEFEF}{\color[HTML]{FE0000} -5.35} & \cellcolor[HTML]{EFEFEF}{\color[HTML]{FE0000} -8.21} & \cellcolor[HTML]{EFEFEF}{\color[HTML]{FE0000} -0.89} & \cellcolor[HTML]{EFEFEF}{\color[HTML]{FE0000} -0.67} \\ \hline
Grok-4.3 & Web-search & 86.09 & 80.23 & 82.45 & 88.51 & 83.04 &  & 83.1 & 81.76 & 77.02 & 92.49 & 81.65 & 83.8 \\
 & ~~~+Evidences & 89.83 & 83.43 & 83.46 & 90.4 & 86.57 & \textbf{} & 84.38 & 77.29 & 78.81 & 91.56 & 82.96 & 85.77 \\
 & \cellcolor[HTML]{EFEFEF}$\Delta$ & \cellcolor[HTML]{EFEFEF}{\color[HTML]{009901} 3.74} & \cellcolor[HTML]{EFEFEF}{\color[HTML]{009901} 3.2} & \cellcolor[HTML]{EFEFEF}{\color[HTML]{009901} 1.01} & \cellcolor[HTML]{EFEFEF}{\color[HTML]{009901} 1.89} & \cellcolor[HTML]{EFEFEF}{\color[HTML]{009901} 3.53} & \cellcolor[HTML]{EFEFEF} & \cellcolor[HTML]{EFEFEF}{\color[HTML]{009901} 1.28} & \cellcolor[HTML]{EFEFEF}{\color[HTML]{FE0000} -4.47} & \cellcolor[HTML]{EFEFEF}{\color[HTML]{009901} 1.79} & \cellcolor[HTML]{EFEFEF}{\color[HTML]{FE0000} -0.93} & \cellcolor[HTML]{EFEFEF}{\color[HTML]{009901} 1.31} & \cellcolor[HTML]{EFEFEF}{\color[HTML]{009901} 1.97} \\ \hline
\end{tabular}
}
\caption{Domain- and language-level macro-F1 scores with evidence-guided web verification. For each model, the table compares an unguided web search with a web search augmented by evidence URLs cited in helpful Community Notes. Scores are reported separately by domain and language, with the final column showing the overall macro-F1. The $\Delta$ rows give the percentage-point change from adding evidence URLs, computed as +Evidence URLs - Web-search. \textcolor{green}{Green} values indicate improvements and \textcolor{red}{red} values indicate declines.}
\label{tab:cf_evigui_scores}
\end{table*}

\begin{table*}[!h]
\resizebox{\linewidth}{!}{
\begin{tabular}{lcccccccccccc}
\hline
\multirow{2}{*}{Model} & \multicolumn{5}{c}{Politics} &  & \multicolumn{5}{c}{Finance} & \multirow{2}{*}{Overall} \\ \cline{2-6} \cline{8-12}
 & en & es & fr & ja & pt &  & en & es & fr & ja & pt &  \\ \hline
Human & 2.0(100) & 1.7(100) & 2.0(99.6) & 2.5(100) & 1.6(100) &  & 1.6(100) & 1.4(100) & 1.7(100) & 1.9(100) & 1.6(100) & 1.9(99.9) \\ \hline
GPT-5-nano & 0.9(48.5) & 1.0(51.7) & 1.1(54.8) & 1.1(59.8) & 1.0(51.3) &  & 1.0(54.1) & 1.0(50.7) & 1.2(61.3) & 1.0(54.3) & 0.9(45.9) & 1.0(53.1) \\
~~+Evidence & 1.4(74.5) & 1.3(70.9) & 1.3(69.8) & 1.8(81.8) & 1.3(67.1) &  & 1.3(73.5) & 1.4(80.6) & 1.3(70.8) & 1.6(81.1) & 1.1(68.1) & 1.4(73.1) \\ \hline
Gemini-2.5-Flash & 3.6(100) & 3.6(99.8) & 4.0(100) & 3.9(99.4) & 3.8(99.6) &  & 3.5(97.4) & 4.1(99.3) & 4.2(100) & 3.8(99.2) & 4.1(99.5) & 3.8(99.4) \\
~~+Evidence & 2.4(76.0) & 2.6(78.6) & 2.5(74.4) & 2.9(77.6) & 2.2(69.1) &  & 2.5(76.0) & 2.4(74.6) & 2.8(79.2) & 3.0(83.5) & 2.3(70.0) & 2.5(75.4) \\ \hline
Grok-4.3 & 19.8(99.8) & 24.1(99.8) & 21.6(99.0) & 21.7(99.4) & 23.2(99.8) &  & 21.0(97) & 23.5(99.3) & 23.5(100) & 21.4(99.2) & 21.2(97.1) & 21.9(99) \\
~~+Evidence & 12.6(98.4) & 15.5(99.2) & 12.8(97.0) & 15.0(98.8) & 13.3(98.0) &  & 14.7(98.8) & 14.4(99.3) & 15.5(99.1) & 12.5(100.0) & 12.5(96.6) & 13.9(98.4) \\ \hline
\end{tabular}}
\caption{Citation statistics for web-search agents. Each cell reports the average number of model-emitted web-source URLs per prediction, with the percentage of predictions containing at least one web-source URL shown in parentheses.}
\label{tab:cf_cite_stats}
\end{table*}

\onecolumn
\clearpage

\begin{tcolorbox}[
  enhanced,
  breakable,
  colback=blue!3,
  colframe=blue!60!black,
  title={\textbf{Domain Classification Prompt}},
  fonttitle=\bfseries,
  boxrule=0.8pt,
  arc=4pt
]
You classify community notes into one domain.\\

\#\#\# TASK\\
Assign exactly one label: Politics, Finance, or Other.\\

\#\#\# DEFINITIONS\\
Politics: Government, public policy, laws/regulation, courts, elections, political figures (in official roles), geopolitics, etc.\\

Finance: Markets (stocks, crypto), banking, business/earnings, personal finance (income, taxes, debt, inflation), gambling/betting, economic indicators (GDP, unemployment, interest rates, trade), etc.\\

Other: Anything not primarily Politics or Finance.\\

\#\#\# EXAMPLES\\
"The Fed raised interest rates to fight inflation." → Finance\\
"Congress passed a bill capping insulin prices." → Politics\\
"Elon Musk's net worth dropped \$10B after the tweet." → Finance\\
"Brazil legalized sports betting; operators now pay 12\% tax." → Politics\\
"The WHO declared mpox a global health emergency." → Other\\
"NASA confirmed water ice on the Moon's south pole." → Other\\
"Lionel Messi won the 2022 FIFA World Cup Golden Ball." → Other\\

\#\#\# OUTPUT\\
Return only one label: Politics | Finance | Other\\

Input: [note]\\
Output:
\end{tcolorbox}
\begin{figure}[h]
\vspace{-\baselineskip}
\caption{Domain classification prompt. This classifies each helpful Community Note into \textsc{Politics}, \textsc{Finance}, or \textsc{Other}.}
\label{fig:cf_domain_class_prompt}
\vspace{-\baselineskip}
\end{figure}

\begin{tcolorbox}[
  enhanced,
  breakable,
  colback=teal!3,
  colframe=teal!80!black,
  title={\textbf{Claim-label Extraction Prompt}},
  fonttitle=\bfseries,
  boxrule=0.8pt,
  arc=4pt
]
You are building a rigorous misinformation detection dataset. Your task is to extract checkworthy factual claims from the provided Tweet and Community Note. Labels are relative to the Community Note (Currently Rated Helpful): "false" means the Tweet asserted the claim and the Note refutes it; "true" means the Note asserts the claim as fact. Do not use external knowledge or infer facts from URLs.\\

\#\# Extraction \& Labeling Rules\\
1. Tweet Claims: Extract Tweet claims only when the Note clearly addresses them. Label as "false" when the Note refutes the Tweet claim; label as "true" only when the Note clearly supports or confirms the Tweet claim. Otherwise drop.\\
2. Note Claims: Label as "true" only when the Note itself makes a factual correction or factual assertion that is directly checkworthy.\\
3. Paraphrase: Rewrite all claims in your own words. Do not use exact quotes. Claims must assert verifiable facts about the world — never about the tweet, note, or source itself. Drop any claim whose entire content is about the credibility, accuracy, sourcing, or wording of the tweet/note rather than the underlying facts.\\
4. Checkworthiness Scope: Keep only substantive high quality real-world facts/events. Drop metadata/status claims whose main point is who posted something, whether an account/source is official, verified, affiliated, fake, or impersonating, or whether a link/livestream/post/page is available, ended, deleted, or reposted. Do not extract these claims even when they are the main correction. Keep an attributed claim only when it states a substantive real-world action, policy, event, filing, speech, decision, date, place, person, organization, or figure independent of account/link/post status. If only metadata/status claims remain, return [].\\
5. Text-Only: The downstream dataset will not include attached images, videos, screenshots, audio, or other media. Drop claims that require viewing, identifying, or interpreting media, including what media shows, who appears in it, where/when it was recorded, or whether it is edited, cropped, staged, AI-generated, or mislabeled.\\
6. Label Precision: "false" = the Tweet asserted it, and the Note refutes it. "true" = the Note asserts it as fact. For substantive attributions, label the attribution itself: if a tweet says "X claims Y", label it "true" if the Note establishes that X made that substantive claim, even if Y is false. Do not apply this attribution rule to account/source authenticity, posting status, link status, or media-identification claims.\\
7. Specificity: Every claim must name the specific people, organizations, countries, jurisdictions, dates, places, figures, metrics, and issuers needed to identify the fact. Do not drop key qualifiers; for example, write "U.S. visas" rather than "visas" when the country matters. Do not add extra details unless they are necessary to disambiguate the claim.\\
8. Language Match: Write each claim in the language of the input text segment.\\

\#\# Density \& Balancing Rules\\
1. Information Density: Do NOT hyper-fragment the text. Merge closely related contextual details such as who, what, when, and where into a single, comprehensive claim.\\
2. Compactness: Keep each claim short while still being standalone and unambiguous. Do not add background, explanation, evidence, rhetorical framing, or provenance unless it is essential to the fact being checked.\\
3. Length Neutrality: Write true and false claims at the same level of detail. Do not make true claims longer by including extra evidence, caveats, or Note explanations.\\
4. Claim Balance: Extract one strongest "false" Tweet claim and one strongest "true" Note claim when both are valid.\\
5. No Forced Claims: If no claim satisfies all rules, return [].\\
6. No Duplicative Opposites: Do not output duplicate claims that merely restate the same fact with opposite labels. A true corrective claim may contradict a false Tweet claim when it adds distinct factual content.\\
7. Output Limit: Strictly limit your output to a few high-value, information-dense claims.\\

\#\# The "Zero-Context Formula" (CRITICAL)\\
Every claim must be 100\% standalone. A reader with zero knowledge of the tweet, note, media, URL, or other claims must fully understand it and fact-check it.\\
- No back-references to other claims: If two claims share a subject, repeat the full subject explicitly in both. Never use "The [noun]" to refer to something established in a prior claim.\\
- No generic subjects: Never use "a photo", "the video", "the account", "the claim", "the note", or similar vaguely in the claim. Inject specific names and entities directly.\\
- No missing qualifiers: Do not omit the country, jurisdiction, organization, issuer, date, place, figure, or metric needed to make the claim identifiable.\\
- No explanatory padding: Do not include why the claim matters, how it was verified, or what the Note is doing.\\
- Structure: [Specific Subject + Explicit Names] + [Event/Context] + [Factual Assertion]\\

\#\# Output Format\\
Return a strict JSON list only. If no claim satisfies all rules, return [].\\
$[$
  \{\\
    "claim": "<paraphrased, dense, explicitly detailed, zero-context factual assertion>",\\
    "label": "true" or "false"\\
  \}
$]$ \\

\#\# Example 1\\
Tweet: The rumor says a major bank collapsed on a silver margin call at 2:47 AM December 28. I cannot verify that. What I can verify is more interesting. JPMorgan filed an 8K on December 27 disclosing 4.875 billion dollars in unrealized silver losses. They flipped from 200 million ounces short to 750 million ounces long physical. The largest position reversal in the history of the silver market happened in the last 30 days and nobody on financial television said a word. The rumor claims 34 billion in emergency Fed repos. Official data shows routine operations under 7 billion. Either the data is lagged or the rumor is wrong. But here is what nobody is asking. Why did JPMorgan suddenly need to own three quarters of a billion ounces of physical silver after spending 15 years on the short side. What did they see coming that made them eat a 5 billion dollar loss just to get positioned the other way. The collapse story might be fiction. The position flip is filed with the SEC. One of those facts will matter more in 90 days than the other. Stop chasing the rumor. Start asking why the smartest bank in commodities just switched sides at the worst possible price and seems fine with it. https://t.co/VZD2WlF6Ux\\
Community Note: No 8-K filing disclosing \$4.875 billion in unrealized silver losses or a position reversal was made by JPMorgan Chase on December 27, 2025. The company's most recent 8-K, filed December 8, 2025, concerns a board member's resignation and makes no mention of silver.    https://jpmorganchaseco.gcs-web.com/sec-filings  https://jpmorganchaseco.gcs-web.com/static-files/b5460587-b02f-448d-9c50-2be5374130af\\
Output:\\
$[$
  \{ \\
    "claim": "JPMorgan Chase filed a December 27, 2025 Form 8-K reporting \$4.875 billion in unrealized silver losses.",\\
    "label": "false"\\
  \}, \\
  \{ \\
    "claim": "JPMorgan Chase's December 8, 2025 Form 8-K did not mention silver.",\\
    "label": "true"\\
  \}
$]$ \\

\#\# Example 2\\
Tweet: Welcome to MEXICO https://t.co/example\\
Community Note: The video was recorded at Pierre Elliott Trudeau Airport (YUL) in Montreal, Canada, and shows agricultural workers arriving to participate in the Mexico-Canada Seasonal Agricultural Workers Program.\\
Output:\\
$[]$ \\

--- \\
Tweet: [tweet] \\
Community Note: [note] \\
Output:
\end{tcolorbox}
\begin{figure}[h]
\vspace{-\baselineskip}
\caption{Claim-label extraction prompt used to construct \dscf{}.}
\label{fig:cf_claim_ext_prompt}
\vspace{-\baselineskip}
\end{figure}

\begin{tcolorbox}[
  enhanced,
  breakable,
  colback=violet!3,
  colframe=violet!60!black,
  title={\textbf{Self-refinement Prompt}},
  fonttitle=\bfseries,
  boxrule=0.8pt,
  arc=4pt
]
Audit the extracted JSON against the original Tweet, Community Note, and rules above. Make the smallest necessary changes.\\

Check label correctness, direct support, standalone specificity, claim completeness, scope violations, duplicates, claim diversity, and label balance. Add a claim only if a central checkworthy claim was missed. Remove claims that are unsupported, vague, over-fragmented, metadata/status-based, media-centered, or duplicative.\\

Prefer up to 2 distinct substantive claims, ideally one strong false Tweet claim and one strong true Note claim when both are valid. Do not force both labels.\\

Return [] if no claim satisfies the rules, otherwise:\\
$[$\{"claim": "...", "label": "true"\}$]$
\end{tcolorbox}
\begin{figure}[h]
\vspace{-\baselineskip}
\caption{Self-refinement prompt for claim-label pair auditing.}
\label{fig:cf_self_ref_prompt}
\vspace{-\baselineskip}
\end{figure}

\begin{tcolorbox}[
  enhanced,
  breakable,
  colback=cyan!3,
  colframe=cyan!70!black,
  title={\textbf{Zero-shot Prompt}},
  fonttitle=\bfseries,
  boxrule=0.8pt,
  arc=4pt
]
You are a fact-checking assistant. Your task is to classify the given claim into true or false.\\

Claim: [claim]\\
Timestamp: [timestamp]\\

Answer with the final verdict of "True" or "False" only. Do not include any explanations or additional text.
\end{tcolorbox}
\begin{figure}[h]
\vspace{-\baselineskip}
\caption{Prompting template for zero-shot classification.}
\label{fig:cf_zeroshot_prompt}
\vspace{-\baselineskip}
\end{figure}

\begin{tcolorbox}[
  enhanced,
  breakable,
  colback=orange!3,
  colframe=orange!70!black,
  title={\textbf{Zero-shot Web-search Prompt}},
  fonttitle=\bfseries,
  boxrule=0.8pt,
  arc=4pt
]
You are a fact-checking assistant with access to web-search. Using evidence retrieved from web, classify the given claim as true or false.\\

Claim: [claim] \\
Timestamp: [timestamp] \\

Answer with the final verdict of "True" or "False" only. Do not include any explanations or additional text.
\end{tcolorbox}
\begin{figure}[!h]
\vspace{-\baselineskip}
\caption{Prompting template for zero-shot classification with access to web-search tool.}
\label{fig:cf_websearch_prompt}
\vspace{-\baselineskip}
\end{figure}

\begin{tcolorbox}[
  enhanced,
  breakable,
  colback=red!3,
  colframe=red!60!black,
  title={\textbf{Zero-shot Evidence-guided Web-search Prompt}},
  fonttitle=\bfseries,
  boxrule=0.8pt,
  arc=4pt
]
You are a fact-checking assistant with access to web-search. Using evidence retrieved from web, classify the given claim as true or false. \\

Claim: [claim] \\
Timestamp: [timestamp] \\
Evidence URLs: [evidence\_urls] \\

Prioritize the Evidence URLs when they are relevant; use web-search if they are insufficient, inaccessible, or do not address the claim. \\

Answer with the final verdict of "True" or "False" only. Do not include any explanations or additional text.
\end{tcolorbox}
\begin{figure}[h]
\vspace{-\baselineskip}
\caption{Prompting template for zero-shot classification with evidence-guided web-search.}
\label{fig:cf_evi_websearch_prompt}
\vspace{-\baselineskip}
\end{figure}

\begin{tcolorbox}[
  enhanced,
  breakable,
  colback=green!3,
  colframe=green!40!black,
  title={\textbf{Human Validation Guidelines}},
  fonttitle=\bfseries,
  boxrule=0.8pt,
  arc=4pt
]
You are evaluating an automatically extracted claim from a Tweet--Community Note pair.
The Community Note shown to you was marked \textbf{Currently Rated Helpful} on X, meaning that X's Community Notes system selected it as useful context for the original Tweet.

Your task is to decide whether the extracted claim is a good benchmark example for misinformation detection.
You are \textbf{not} judging whether the original Tweet is well written, whether the Community Note is persuasive, or whether you personally agree with either one.
Instead, compare the extracted claim with the original Tweet, the helpful Community Note, the assigned domain, the factuality label, the label timestamp, and the URLs cited in the Note.

The extracted claim may differ from the exact wording of the Tweet or Community Note.
This is allowed.
A claim may paraphrase the Tweet or Note, combine closely related details, or make an implied factual assertion explicit.
However, the claim should not introduce new factual content that is not addressed by the Tweet--Note pair.

The factuality label is \textbf{relative to the helpful Community Note}.
A claim labeled \textbf{False} should be a claim made or implied by the Tweet that the Community Note refutes, contradicts, or corrects.
A claim labeled \textbf{True} should be a factual claim that is asserted, supported, or clarified by the Community Note.
For attribution claims, \textbf{True} means that the person or organization really made the attributed statement, proposal, filing, or action, even if the embedded statement is itself false.

You are also given a timestamp for the factuality label.
Use this timestamp when judging time-sensitive claims.
Some claims may change truth value over time; for example, a claim such as ``Kenya's economy is stable'' may be judged differently in different years.
For Q5, evaluate the claim according to the evidence and real-world context at the given timestamp, not according to what may be true today.

For Q1--Q4, use only the provided Tweet, Community Note, and cited URLs.
Do not use outside knowledge to override the Community Note.
Q3 checks whether the dataset label is consistent with the helpful Community Note.
Q5 is separate: it asks for an independent factuality judgment based on the cited URLs and, only if necessary, your own fact-checking.

Score each dimension independently.
For example, a claim may be self-contained but assigned the wrong label, or correctly labeled but supported by weak evidence URLs.

\vspace{6pt}
\hrule
\vspace{6pt}
\textbf{Claim Quality} 
\vspace{6pt}

\textbf{Q1a --- Self-contained:} Can the claim be understood and evaluated without reading the original Tweet or Community Note? (\textit{Yes/No}) \\ \\
\textbf{Yes} if the claim names the necessary people, organizations, places, dates, quantities, and context. \\
\textbf{No} if the claim relies on vague references such as “this video,” “the post,” “the claim,” “the account,” “he,” “she,” or “they” without enough context.\\

\textbf{Q1b --- Checkable:} Is the claim a specific factual assertion that could in principle be verified/fact-checked? (\textit{Yes/No})\\ \\
\textbf{Yes} for concrete claims about events, actions, dates, numbers, policies, statements, filings, or outcomes. \\
\textbf{No} for opinions, predictions, jokes, vague allegations, commentary, or claims that are too unclear to verify.

\vspace{6pt}
\hrule
\vspace{6pt}
\textbf{Domain Correctness}
\vspace{6pt}

\textbf{Q2:} Does the assigned domain (Politics / Finance) match the claim’s primary subject matter? (\textit{Yes/No})\\ \\
\textbf{Politics:} government, elections, public policy, laws, courts, regulation, geopolitics, political institutions, or public officials acting in official roles.\\
\textbf{Finance:} markets, stocks, crypto, banking, business, earnings, scams involving money, debt, taxes, inflation, interest rates, gambling/betting, or economic indicators.\\
\textbf{Other:} any claim whose main subject is not Politics or Finance.

\vspace{6pt}
\hrule
\vspace{6pt}
\textbf{Label Correctness}
\vspace{6pt}

\textbf{Q3a --- Note Relevance:} Is the Community Note directly relevant to the given claim? (\textit{Yes/No})\\ \\
\textbf{Yes} only if the Community Note addresses the same factual assertion as the extracted claim. \\
\textbf{No} if the note is about the same broad topic but not the same specific claim, or if the claim adds information not addressed by the note.\\

\textbf{Q3b --- Label Accuracy (if Q3a=Yes):} Given what the Community Note says, is the factuality label associated with the claim correct? (\textit{Yes/No}) \\ \\
\textbf{True} means the claim as written is accurate according to the helpful Community Note. For attribution claims, this means the person or organization really did make the attributed statement, proposal, filing, or action, even if the embedded statement is itself false. \\
\textbf{False} means the helpful Community Note refutes, contradicts, or corrects the claim as written. \\
The wording of the claim may differ from the Tweet or Note; judge whether the same underlying factual assertion is addressed. \\
Do not use outside knowledge to override the Community Note. The goal is to check whether the extracted label is consistent with the helpful note.

\vspace{4pt}
\textbf{Examples}

\textit{Example A: Note directly refutes the claim.} \\
Claim: ``Hamas carried out airstrikes in Qatar.'' \\
Community Note: ``The airstrikes were carried out by Israel, not Hamas.'' \\
Q3a: \textbf{Yes}. The note addresses the same factual assertion. \\
If the dataset label is \textbf{False}, Q3b: \textbf{Yes}. \\

\textit{Example B: Claim wording differs, but the same fact is addressed.} \\
Claim: ``JPMorgan Chase filed a December 27, 2025 Form 8-K reporting \$4.875 billion in unrealized silver losses.'' \\
Community Note: ``No such 8-K filing was made by JPMorgan Chase on December 27, 2025.'' \\
Q3a: \textbf{Yes}. The claim is paraphrased, but the note clearly addresses it. \\
If the dataset label is \textbf{False}, Q3b: \textbf{Yes}. \\

\textit{Example C: Note is related but not specific enough.} \\
Claim: ``A new tax law increased income taxes for small businesses in France in 2025.'' \\
Community Note: ``France passed several budget measures in 2025.'' \\
Q3a: \textbf{No}. The note is about the same broad topic, but it does not address the specific tax claim. \\
Skip Q3b. \\

\textit{Example D: Attribution claim.} \\
Claim: ``The Indian finance minister said the government would not raise fuel taxes.'' \\
Community Note: ``In the press conference, the Indian finance minister did say that the government would not raise fuel taxes.'' \\
Q3a: \textbf{Yes}. \\
If the dataset label is \textbf{True}, Q3b: \textbf{Yes}, even if the annotator believes the government later raised taxes. 
The label checks whether the statement was made, not whether the future policy happened.

\vspace{6pt}
\hrule
\vspace{6pt}
\textbf{Evidence URLs Relevance}
\vspace{6pt}

\textbf{Q4:} How well do the extracted URLs from Community Note support or refute this specific claim? (\textit{Direct/Partial/Weak/None}) \\ \\
Direct --- At least one cited URL directly supports or refutes the claim as written, with the same key entities, dates, places, figures, policies, or events. \\
Partial --- The cited URL is relevant and helps evaluate the claim, but one or more important details are missing or require inference. \\
Weak / Tangential --- The cited URL is about the same broad topic, but it does not clearly support or refute the specific claim. \\
None --- No cited URL is provided, the URL is inaccessible, or the cited URL does not address the claim at all.

\vspace{6pt}
\hrule
\vspace{6pt}
\textbf{Expert Audit: Independent Factuality}
\vspace{6pt}

\textbf{Q5:} Based on the cited URLs, and only if necessary your own fact-checking, what is the real-world factuality of the claim? 
(\textit{True/False/Insufficient Evidence}) \\ \\
True --- The claim is supported by reliable evidence. \\
False --- The claim is contradicted by reliable evidence. \\
Insufficient Evidence --- The available evidence is not enough to determine whether the claim is true or false. \\

This question is separate from Q3 which evaluates whether the dataset label is consistent with the helpful Community Note; but this question evaluates the claim's independent real-world factuality.

\end{tcolorbox}
\begin{figure*}[h]
\vspace{-\baselineskip}
\caption{Human validation guidelines for \dscf{}.}
\label{fig:cf_he_guide}
\vspace{-\baselineskip}
\end{figure*}

\end{document}